\documentclass[11pt]{article}

\usepackage{etoolbox}
\usepackage{amsmath}
\providetoggle{blind}
\providetoggle{showtodos}
\providetoggle{splitsections}
\settoggle{blind}{true}
\settoggle{splitsections}{false}
\settoggle{showtodos}{false}

\newcommand{\ourtitle}{%
  Responsiveness Verification:\\
  Will Predictions Change? How Much? How Often?
}
\newcommand{\ourshorttitle}{Responsiveness Verification}

\iftoggle{blind}{
\newcommand{\repourl}[0]{\href{https://anonymous.4open.science/r/rv-iclr2027/}{in our anonymized repository}}
}{
\newcommand{\repourl}[0]{\href{https://github.com/ustunb/reachml}{on GitHub}}
}

\PassOptionsToPackage{numbers,sort&compress}{natbib}
\usepackage[venue=arxiv]{ext/mlconf}
\setcitestyle{numbers,square,comma}

\providetoggle{arxiv}
\providetoggle{neurips}
\providetoggle{icml}
\providetoggle{iclr}
\settoggle{neurips}{false}
\settoggle{arxiv}{true}
\settoggle{iclr}{false}
\settoggle{icml}{false}

\mlauthor[equal, corresponding]{Harry Cheon}{CSE}{University of California, San Diego}{scheon@ucsd.edu}
\mlauthor[equal]{Meredith Stewart}{CSE}{University of California, San Diego}{mds010@ucsd.edu}
\mlauthor{Bogdan Kulynych}{}{Lausanne University Hospital}{bogdan@kulyny.ch}
\mlauthor{Tsui-Wei Weng}{HDSI}{University of California, San Diego}{lweng@ucsd.edu}
\mlauthor[corresponding]{Berk Ustun}{HDSI}{University of California, San Diego}{berk@ucsd.edu}

\usepackage[T1]{fontenc} 
\usepackage{environ,comment}
\usepackage{amsfonts,bbm,bm,courier,nicefrac,microtype}
\usepackage{amsmath,amsthm,amssymb}
\usepackage{array,booktabs,tabularx,multirow,colortbl,hhline}
\usepackage{eqnarray}
\usepackage[inline]{enumitem}\setlist{nosep,leftmargin=*} 
\usepackage{xifthen}%
\usepackage{thmtools,thm-restate}
\usepackage{xspace}
\usepackage{setspace}
\usepackage{tikz}
\usepackage{listings}
\usepackage{seqsplit}
\usepackage[group-separator={,},group-minimum-digits={3}]{siunitx}
\renewcommand{\thmcontinues}[1]{}
\usepackage{inconsolata}
\usepackage{pifont}
\usepackage{accents}

\usepackage{graphicx,xcolor,placeins}
\usepackage{wrapfig}
\usepackage{caption}

\definecolor{good}{HTML}{BAFFCD}
\definecolor{bad}{HTML}{FFC8BA}

\usepackage{hyperref}
\hypersetup{colorlinks=true, urlcolor=blue, linkcolor=blue, citecolor=blue, linkbordercolor=blue, pdfborderstyle={/S/U/W 1}}

\renewcommand{\arraystretch}{1.2}
\newcolumntype{H}{>{\setbox0=\hbox\bgroup}c<{\egroup}@{}}
\newcolumntype{L}[1]{>{\raggedright\arraybackslash}p{#1}}
\newcolumntype{R}[1]{>{\raggedleft\arraybackslash}p{#1}}
\newcommand{\textheader}[1]{{\bfseries{#1}}}
\newcommand{\cell}[2]{\setlength{\tabcolsep}{0pt}\begin{tabular}{#1}#2 \end{tabular}}

\usepackage{natbib}

\setitemize{leftmargin=1em,topsep=0.1em,parsep=0.5pt,}
\setlist[enumerate]{leftmargin=*, label= {\arabic*.}, itemsep=0.5em}

\newlist{bulletlist}{itemize}{1}
\setlist[bulletlist]{label=\textbullet,itemsep=0.2em,leftmargin=1em,topsep=0.1em,parsep=0.5pt}

\newlist{plainlist}{itemize}{1}
\setlist[plainlist]{label={},itemsep=0.2em,leftmargin=0em,topsep=0.1em,parsep=0.5pt}

\usepackage{ext/prettythm}

\usepackage{algorithm}
\usepackage[noend]{algpseudocode}
\newcommand{\algrule}[2]{\par\noindent\hfil\rule{#1}{#2}\hfil\par\vskip.25\baselineskip}

\algnewcommand{\alginput}[2]{\Statex{Input:~#1}\Comment{#2}}
\algrenewcommand\algorithmicrequire{\textbf{Input:}}
\algrenewcommand\algorithmicensure{\textbf{Output:}}
\algnewcommand\algorithmicinput{\textbf{Input}}
\algnewcommand\algorithmicinitialize{\textbf{Initialize}}
\algnewcommand\algorithmicbigstep{\textbf{Step}}
\algnewcommand\INPUT{\item[\algorithmicinput]}
\algnewcommand\INITIALIZE{\item[\algorithmicinitialize]}
\algnewcommand{\STEP}[1]{\item[\algorithmicbigstep]{\textbf{#1}}}
\algnewcommand{\InputExplanation}[2][.6\linewidth]{\leavevmode\hfill\makebox[#1][r]{~{\footnotesize{#2}}}}
\algnewcommand{\InitializationExplanation}[2][.6\linewidth]{\leavevmode\hfill\makebox[#1][r]{~{\footnotesize{#2}}}}
\algnewcommand{\StateComment}[2]{\State{#1}\InputExplanation{#2}}
\algnewcommand{\alginitialize}[2]{\Statex{#1}\InitializationExplanation{#2}}
\algrenewcommand\algorithmiccomment[2][]{#1\hfill\textit{\scriptsize{#2}}}

\usepackage[capitalise]{cleveref}
\Crefformat{figure}{Fig.~#2#1#3}
\crefname{boxtheorem}{Thm.}{Thms.}
\crefname{boxlemma}{Lemma}{Lemmas}
\crefname{boxcorollary}{Corollary}{Corollaries}
\crefname{boxproposition}{Prop.}{Props.}
\crefname{boxdefinition}{Def.}{Defs.}
\crefname{boxremark}{Remark}{Remarks}
\crefname{boxassumption}{Assumption}{Assumptions}
\crefname{boxexample}{Example}{Examples}
\Crefformat{boxdefinition}{Def.~#2#1#3}
\Crefformat{boxproposition}{Prop.~#2#1#3}
\Crefformat{boxtheorem}{Thm.~#2#1#3}

\makeatletter
\newcommand*{\centerfloat}{%
  \parindent \z@
  \leftskip \z@ \@plus 1fil \@minus \textwidth
  \rightskip\leftskip
  \parfillskip \z@skip}
\makeatother

\newcommand{\textds}{\texttt}
\newcommand{\textfn}[1]{\texttt{#1}}

\newcommand{\indic}[1]{\mathbbm{I}[#1]}

\newcommand{\prob}[1]{\textnormal{Pr}\big(#1\big)}
\renewcommand{\Pr}[1]{\textnormal{Pr}(#1)}
\newcommand{\E}[0]{\mathbb{E}}
\newcommand{\B}{\{0,1\}}
\newcommand{\R}{\mathbb{R}}
\newcommand{\Z}{\mathbb{Z}}

\newcommand{\optpar}[2]{\ifthenelse{\isempty{#2}}{#1}{#1({#2})}}

\newcommand{\vecvar}[1]{\bm{#1}}

\newcommand{\xb}{\vecvar{x}}
\newcommand{\X}{\mathcal{X}}
\newcommand{\Y}{\mathcal{Y}}

\newcommand{\clf}[1]{\optpar{f}{#1}}

\newcommand{\xp}{\xb}
\newcommand{\xq}{\xb'}

\newcommand{\st}{\textnormal{s.t.}}

\newcommand{\mipwhat}[1]{\emph{\scriptsize #1}}
\newcommand{\ajpos}{a_{j}^+}
\newcommand{\ajneg}{a_{j}^-}

\newcommand{\intrange}[1]{[#1]}

\newcommand{\Rhat}[0]{{\smash{\hat{X}}}}
\newcommand{\respdist}[1]{P_{#1}}

\newcommand{\impscore}[3]{%
\ifthenelse{\isempty{#3}}{%
\phi_{{#2}}({#1})%
}{%
\phi_{{#2}}({#1};{#3})%
}%
}

\newcommand{\impscorevec}[2]{%
\ifthenelse{\isempty{#2}}{%
\bm{\phi}({#1})%
}{%
\bm{\phi}({#1};{#2})%
}%
}

\newcommand{\LR}{{$\mathsf{LR}$}}

\newcommand{\XGB}{{$\mathsf{XGB}$}}

\definecolor{bad}{HTML}{ff8080}
\definecolor{good}{HTML}{BAFFCD}
\definecolor{recourse}{HTML}{8FAADC}

\algnewcommand\algorithmicforeach{\textbf{for each}}
\algdef{S}[FOR]{ForEach}[1]{\algorithmicforeach\ #1\ \algorithmicdo}

\newcommand{\runningrset}[0]{\Rhat{}}

\newcommand{\rejector}[1]{\mathsf{Rejector}{(#1)}}

\newcommand{\respfunc}[0]{\bm{\varepsilon}}
\newcommand{\downstream}[0]{\varepsilon}

\newcommand{\ytarget}[1]{\smash{\hat{Y}_{#1}}}

\newcommand{\AlgNoIndent}[1]{%
  \Statex \hspace{-\algorithmicindent}#1%
}

\newcommand{\ab}[0]{\vecvar{a}}

\newcommand{\Aset}[1]{\optpar{A}{#1}}

\newcommand{\Ajset}[1]{A_j(#1)}

\newcommand{\Imodel}[1]{\optpar{I}{#1}}

\newcommand{\rsptrue}[1]{\smash{\optpar{\rho}{#1}}}
\newcommand{\rsphat}[1]{\smash{\optpar{\hat{\rho}_n}{#1}}}
\newcommand{\rsphata}[1]{\smash{\optpar{\hat{\rho}}{#1}}}
\newcommand{\sampResp}[1]{\smash{\optpar{\hat{\rho}_n}{#1}}}
\newcommand{\rsphatmax}[2]{\smash{\bar{\rho}_{#2}}}

\newcommand{\rspub}[1]{\smash{\optpar{\rho^{\text{all}}}{#1}}}
\newcommand{\rsplb}[1]{\smash{\optpar{\rho^{\text{any}}}{#1}}}

\newcommand{\reachdist}[1]{P^\mathrm{\,reach}_{\xp}}
\newcommand{\sampReach}[1]{\smash{\optpar{\hat{X}_n}{#1}}}

\newcommand{\betaQuant}[3]{\mathsf{B}_{#1}({#2}, {#3})}

\newcommand{\ci}{\smash{C}}
\newcommand{\nmin}[1]{\smash{N^{\text{min}}}({#1})}
\newcommand{\altP}[0]{\delta}

\newcommand{\partition}{\mathcal{J}}

\newcommand{\intFamily}{\mathcal{I}}

\newlist{constraints}{enumerate}{3}
\setlist[constraints]{label={\arabic*.},leftmargin=*}

\newcommand{\prt}{S}

\newcommand{\thresh}{\tau}

\newcommand{\relAset}[1]{\Aset{}_\text{rel}}
\newcommand{\origAset}[1]{\Aset{\xp}}

\newcommand{\nSucc}[0]{K}

\newcommand{\scope}[1]{\mathsf{scope}(#1)}

\newcommand{\interub}[2]{\bar{A}_{#1}(#2)}
\newcommand{\interlb}[2]{\underaccent\bar{A}_{#1}(#2)}

\normalmarginpar
\iftoggle{showtodos}{%
\newcommand{\tbu}[1]{{\todo[]{[Berk: {#1}]}}}
\newcommand{\lily}[1]{\todo[]{[Lily: {#1}]}}
\newcommand{\bogdan}[1]{\todo[]{[Bogdan: {#1}]}}
}{
\newcommand{\tbu}[1]{}
\newcommand{\lily}[1]{}
\newcommand{\bogdan}[1]{}
}
\presetkeys{todonotes}{size=\scriptsize, color=green!30}{}

\begin{document}
\mlmaketitle{\ourtitle}
\begin{abstract}
Machine learning models are often used in applications where their inputs change due to routine interactions, strategic manipulation, or noise. In such settings, models can undermine safety as these changes can lead them to predict over regions of the input space they have not seen. We propose to address these safety risks by measuring responsiveness---the probability that a model output attains a target prediction under feasible interactions defined by the interaction model. We develop algorithms to estimate responsiveness for any machine learning model, and a framework to specify broad classes of interaction models. We pair these algorithms with statistical guarantees that support practical validation. We demonstrate how our tools can promote safety and reliability across domains by detecting preclusion in recidivism prediction, estimating the cost of gaming in content moderation, and testing the robustness of benchmarks for LLMs.
\end{abstract}

\iftoggle{arxiv}{
\vspace{0.25em}
{}}

\section{Introduction}
\label{Sec::Introduction}

\definecolor{fail}{HTML}{ff8080}
\definecolor{good}{HTML}{BAFFCD}
\definecolor{target}{HTML}{f5d16f}

Machine learning models are often used in tasks where their inputs change -- be it due to user interactions~\citep[][]{hardt2016strategic,miller2020strategic,perdomo2020performative}, stress tests~\citep[][]{kusner2017counterfactual,veitch2021counterfactual}, distributional shifts~\citep[][]{goodfellow2014explaining,madry2017towards,miller2021accuracy}, or noise~\cite[][]{xia2020part,tzeng2025improving}. In such settings, a model that is accurate on held-out data can still undermine safety when its outputs respond to changes in its inputs in ways we fail to anticipate:
\begin{itemize}[itemsep=0.2em]
    \item \emph{Content Moderation:} Consider an online platform that uses a machine learning model to detect bot accounts at scale~\citep{shao2018spread,milli2024engagement}. Here, bot accounts may be able to evade detection by manipulating their profile or behavior (e.g., reducing followers, or posting more often). %
    
    \item \emph{Lending:} Consider a lender who uses a machine learning model to underwrite loans. Here, applicants who are denied may be permanently denied access to credit as their prediction may not change under any possible action they can take~\citep{ustun2019actionable, kothari2024prediction}. %
    
    \item \emph{Benchmarking:} Consider an online retailer adopting a language model to streamline customer support based on its performance in a safety benchmark. Here, the retailer may adopt a different model, or none, if the benchmark scores and rankings change as a result of benign changes to the prompt~\citep{xie2025sorry} or context~\citep{wang2025inadequacy}.%
\end{itemize}
These failure modes arise because machine learning algorithms are designed to learn \emph{pointwise mappings}. Simply put, we can expect an accurate model to correctly map inputs to their corresponding output, but we cannot expect a provably correct change in outputs when its inputs change.

Existing approaches have sought to mitigate failure modes by developing algorithms to learn models whose predictions will change in predictable ways~\cite[e.g.,][]{madry2017towards,perdomo2020performative,ross2021learning}. This approach is challenging because we often lack the labels that give the correct prediction after an input changes, the machinery to specify how inputs can change, and the algorithms to enforce these specifications during training~\citep[][]{barocas2020hidden,venkatasubramanian2020philosophical}. In effect, many interactions must not only adhere to semantic constraints (e.g., only changing in one direction) but also trigger downstream effects. In practice, these challenges have meant that many of the tools to build models that operate safely under recourse, gaming, and performativity have often sought to achieve safety by compromising fidelity. On the one hand, researchers have sought to design algorithms to learn models that will behave in predictable ways under simple interaction models. On the other, practitioners can resort to extreme measures, such as dropping all mutable features, even when those features contain useful signal. 

In this paper, we present an alternative approach: instead of learning models that will behave in predictable ways when their inputs change, we treat models as given and test how their outputs change under interactions. 
In contrast to learning, testing allows us to work with richer interaction models and can be used at multiple points in the model lifecycle. We capitalize on these advantages by developing theory, methods, and software that make testing rigorous, versatile, and practical. We introduce a foundational notion of \emph{responsiveness} that applies across machine learning models and interaction models, supports decisions, and helps anticipate and mitigate harms. Our main contributions include:
\begin{enumerate}[itemsep=0.25em,topsep=0.2em]
\label{List::Contributions}

\item We introduce a new approach for model validation in which we measure the responsiveness of its predictions. Our approach can ensure that models operate safely by detecting if, how, and when predictions change under realistic interaction models.

\item We develop machinery for defining feasible interactions through actionability constraints, and downstream effects. Our machinery allows practitioners to faithfully model interactions and to evaluate their specification through sensitivity analysis and ablation studies.

\item We design fast algorithms to estimate the responsiveness of predictions for any model, any dataset, and any interaction model. We pair our estimates with formal guarantees that can guide practical decisions in test design 
and yield interpretable claims on responsiveness.

\item We demonstrate how responsiveness verification can promote safety across domains, including recidivism prediction, content moderation, and benchmarking LLMs. Our results highlight how responsiveness can be used to detect and mitigate failure modes across domains.

\item We provide a Python library for responsiveness verification \repourl{}.

\end{enumerate}

\begin{figure*}[t]
    \centering
\includegraphics[
    trim=0.11in 2.8in 0.5in 0.0in,
    clip,%
    width=0.95\linewidth,%
    page=1
    ]{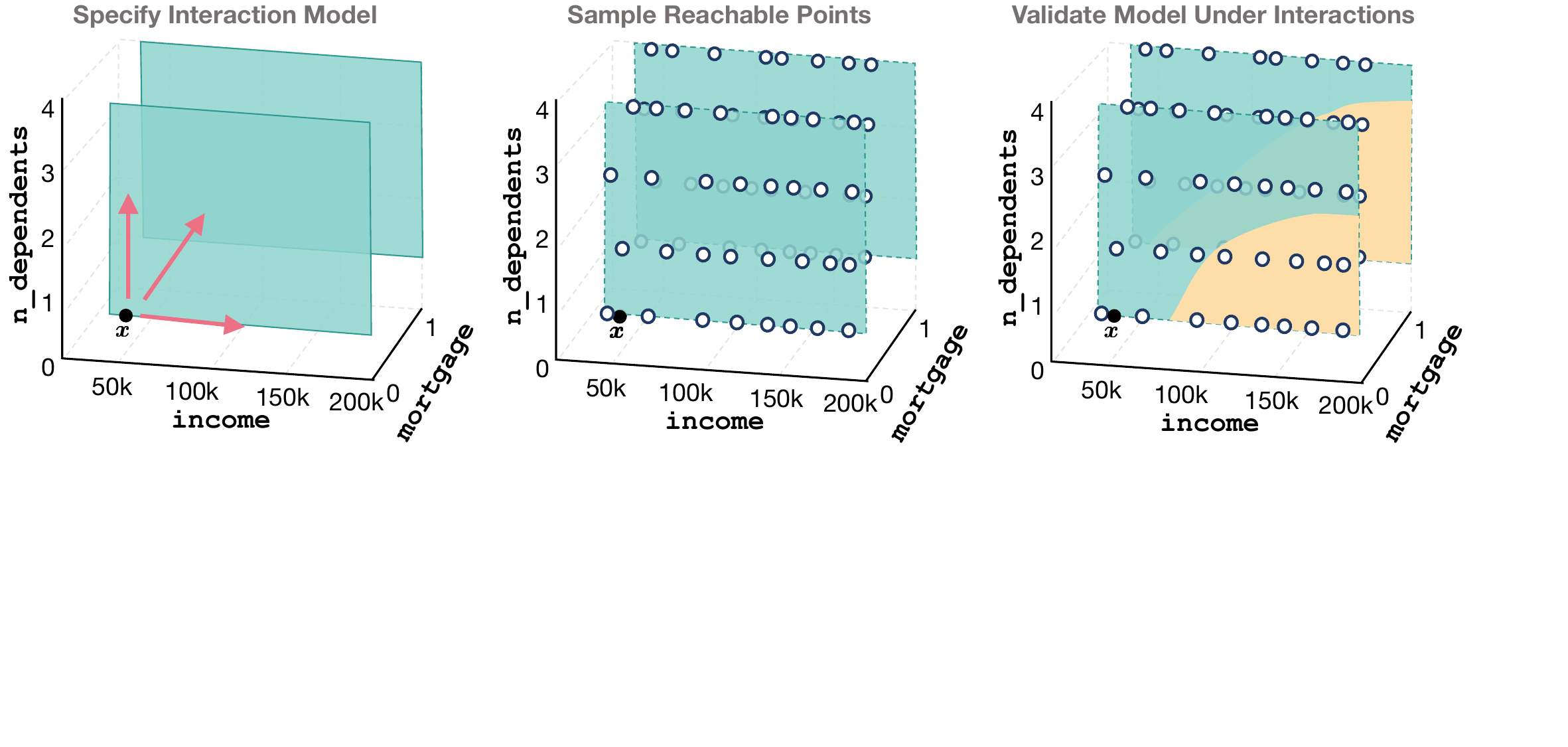}
    \caption{
    Detecting safety violations in models by measuring responsiveness. 
    Safety violations arise when model outputs are overly or insufficiently responsive to interactions.
    On the right, we consider detecting preclusion of a model with one discrete (\textfn{n\_dependents}), one continuous (\textfn{income}), and one binary (\textfn{mortgage}) feature. \textfn{mortgage} and \textfn{n\_dependents} can only increase.
}
    \label{Fig::Overview}
    \vspace{-1em}
\end{figure*}

\paragraph{Related Work}
\label{Sec::RelatedWork}
\todo[inline,caption={}]{Restructure: \begin{itemize}
    \item Safety issues (our work is motivated by [insert list]
    \item Recourse + counterfactual invariance as falsification 
    \item We are most technically related to recourse, insert most methods list

    P1: 
    - Our work deals with these safety violations
    - Other works have dealt with this by:
        - learning models with resp. guarantees
        - falsification

    P2: How do we deal with this? By testing. Foundational framework on two fronts
    - Machinery to specify complex interactions with constraints and downstream effects (versatile, resolves underspecification).
    - Statistical guarantees / certification

    P3: Recourse verification compare and contrast
\end{itemize}}
Our work is motivated by a broad class of safety violations that stem from interactions with the inputs of a model. This class captures a wide range of effects studied in areas such as adversarial robustness~\citep{goodfellow2014explaining,madry2017towards}, strategic classification~\citep{hardt2016strategic, dong2018strategic}, performative prediction~\citep{perdomo2020performative,mendler2020stochastic}, counterfactual invariance~\citep{veitch2021counterfactual,kusner2017counterfactual}, and algorithmic recourse~\citep{ustun2019actionable,pawelczyk2023robust,cheon2025responsiveness}.

Much of the work in these areas focuses on learning models that are robust to a specific effect -- e.g., to gaming in strategic classification~\citep{hardt2016strategic, dong2018strategic, chen2020learning, levanon2021strategic}, or to imperceptible perturbations in adversarial robustness~\citep{madry2017towards, zhang2019theoretically, Wong2018Provable, zhang2020crownibp}. We propose to address these violations through testing. Testing only needs to check how a fixed model responds to interactions, and can thus accommodate far richer interaction models than we could enforce during training.

On the technical front, our proposal is most closely related to research on algorithmic recourse. Our machinery builds on the model-agnostic approach to recourse verification of \citet{kothari2024prediction}, who construct the set of all points reachable through interactions. We generalize their approach to any dataset and replace enumeration with sampling. Sampling makes verification computationally viable for complex interaction models with probabilistic downstream effects. As a result, our approach can verify recourse for a broad class of problems. Existing methods find recourse actions~\citep[see e.g.,][]{wachter2017counterfactual, mothilal2020explaining, karimi2021algorithmic}, often for specific model classes~\citep{ustun2019actionable, mothilal2020explaining}, or certify the feasibility of recourse only over discrete feature spaces~\citep{kothari2024prediction}. In contrast, our approach can certify the feasibility of recourse over any dataset, requires only black-box access, and returns practical guarantees. %

\section{Framework}
\label{Sec::Framework}%

Our goal is to study how the output of a model changes due to interactions on its inputs.
We consider a model $\clf{}: \X \to \Y$ that outputs an outcome $y \in \Y$ from a set of $d$ \emph{features} $\xp = [x_{1}, \ldots, x_{d}] \in \X$.
We assume that features encode semantically meaningful characteristics for a task at hand (e.g., \textfn{income} as opposed to pixel intensities or token embeddings). 
We also assume that we may only have \emph{black-box} access to the model---i.e., we can query its output for arbitrary inputs $\xp$, but may not have its parameters.
We consider a subset of instances $\xp \in \X$ that satisfy specific conditions (e.g., $\clf{\xp} = 0$) for our validation tasks.

\todo[inline]{
- Understanding how predictions change from interactions over semantically meaningful features is hard because:
   - constraints
   - downstream effects
   follow up with examples for each.
- We present math model / machinery. It doesn't make any assumptions (constraints and downstream effects are specified, not assumed).
Update definition of I(x) to be one big definition environment with A(x) and P(x).

Update table to show how we use rho (x); Remove domain
}

Understanding how outputs change from interactions over semantically meaningful features is difficult as interactions operate in a constrained space (e.g., an ordinal encoding of \texttt{education\_level} can only increase). Interactions may also contain indirect changes to features (e.g., increasing \texttt{num\_posts} may increase \texttt{url\_count}). Below, we present a formal framework for modeling interactions:

\begin{definition}[Interaction Model]\normalfont
\label{Def::InteractionModel}
Given an individual with features $\xp \in \X$, the interaction model is $\Imodel{\xp} = (\Aset{\xp}, \respdist{\xp,\cdot})$, where:
\begin{bulletlist}
    \item Action set \(\Aset{\xp} \subseteq \mathbb R^d\) is the set of feasible actions. We define
    \begin{equation*}
    \Aset{\xp} :=
    \Bigl\{
    \ab \in \prod_{j=1}^d \Ajset{\xp}
    :
    G(\xp,\ab) \le 0,\;
    H(\xp, \ab) = 0
    \Bigr\},
    \end{equation*}
    where \(\Ajset{\xp} \subseteq \mathbb R\) encodes feature-level constraints, 
    \(G(\xp,\ab) \le 0\) encodes joint inequality constraints, and 
    \(H(\xp,\ab)=0\) encodes joint equality constraints.

    \item \(\respdist{\xp,\cdot} := \{\respdist{\xp,\ab} : \ab \in \Aset{\xp}\}\) is a family of downstream effect distributions.
    Downstream effect distributions \(\respdist{\xp,\ab}\) may be specified through conditional distributions
    \(
    P_j(\cdot \mid \xp, \ab, \downstream_{\text{pa}(j)}),
    \)
    where \(\operatorname{pa}(j)\) denotes the upstream parents of feature \(j\) among the downstream effects.
\end{bulletlist}
\(\Imodel{\xp}\) induces a reachable point
   \(\xq = \xp + \ab + \respfunc\), with action \(\ab\) and downstream effect \(\respfunc\).
\end{definition}
The interaction model captures how the features of a point can change through two components. The \emph{action set} encodes how we can change the inputs to a model from a given point. We specify it in terms of \emph{actionability constraints} on changes. These include \emph{feature-level} constraints, which restrict each feature independently (e.g., enforcing monotonicity), and \emph{joint} constraints, which restrict multiple features at the same time (e.g., preserving a one-hot encoding). The \emph{downstream effect distribution} encodes what can change as a result of a feasible action, i.e., one that belongs to the action set. It can be specified to capture a variety of post-action effects, from \emph{random effects} (e.g., measurement noise) to \emph{causal effects} (e.g., increasing \texttt{num\_posts} may increase \texttt{url\_count}).

We use the interaction model to measure how the output of $\clf{}$ responds to interactions at $\xp$:
\begin{definition}[Responsiveness]\normalfont
\label{Def::Responsiveness}
    Given a model $\clf{}$, point $\xp$, interaction model $\Imodel{\xp}$, and a set of target outputs $\ytarget{\xp} \subseteq \Y$, let $\xq$ denote a reachable point induced by $\Imodel{\xp}$.
    The \emph{responsiveness} of $\clf{}$ at $\xp$ is:
    \begin{align*}
        \rsptrue{\xp} := \mathbb{E}_{\xq}\Bigl[{\mathbb{I}\bigl[\clf{\xq} \in \ytarget{\xp}}\bigr]\Bigr],
    \end{align*}
     the probability that an interaction attains a target output.
\end{definition}

We can use responsiveness in various tasks and applications (see \cref{Tab::RespTasks}).
When the space of feasible actions is simple (i.e., we can enumerate through all of them), answering these questions is trivial~\citep{kothari2024prediction}.
However, enumeration is either infeasible (e.g., feature space is not discrete) or computationally expensive. 
In what follows, we present a Monte Carlo sampling procedure to estimate responsiveness and pair it with statistical guarantees in \cref{Sec::InferenceGuarantees}.

\begin{table}[t]
   \centering
   \resizebox{\linewidth}{!}{
   \begin{tabular}{L{0.38\linewidth}llL{0.4\linewidth}}
          \textbf{Validation Task} & 
        \textbf{Responsiveness Semantics} &
        \textbf{Application} &
        \textbf{Interaction Model} 
        \\
        \midrule
        \cell{l}{\textbf{Falsification}: Does the output\\ change under interactions?}
        & \cell{c}{
        \( \rho^\text{max}(\xp) = \max_{\xq} \indic{\clf{\xq} \in \ytarget{\xp}}\)
        } 
        & Falsify Preclusion in Lending
        & \cell{l}{
        \(a_{\texttt{late\_payments}} \geq 0\) \\
        \(a_{\texttt{checking\_bal}} \geq 0\) \\
        \(\varepsilon_{\texttt{savings\_bal}}
            \sim P(\cdot \mid x, a_{\texttt{checking\_bal}})\)
        }
        \\
        
        \midrule
        
        \cell{l}{\textbf{Estimation}: How often do outputs\\ change under interactions?}
        & \cell{l}{\(\sampResp{\xp} := \frac{1}{n} \sum_{i=1}^n \mathbb{I}\left[ \clf{\xq_i} \in \ytarget{\xp} \right] \)}
        & \cell{l}{Measure Prevalence of Gaming\\ in Content Moderation}
        & \cell{l}{
        \(a_{\texttt{num\_tweets}} \geq 0\) \\
        \(a_{\texttt{followers}} = 0\) \\
        \(\varepsilon_{\texttt{url\_count}}
            \sim P(\cdot \mid a_{\texttt{num\_tweets}})\)
        }
        \\
        
        \midrule
        
        \cell{l}{\textbf{Testing}: Is an output sufficiently\\ responsive under interactions?}
        & \cell{l}{\( H_0:\rsptrue{\xp} \geq \thresh, \; H_1: \rsptrue{\xp} < \thresh \)}
        & \cell{l}{Testing Performance Invariance in\\ Safety Benchmarks for LLMs}
        & \cell{l}{
        \(a \in \{\texttt{misspell},
        \texttt{roleplay},
        \texttt{translate}\}\) \\
        \(\varepsilon \sim P_{\texttt{decode}}(\cdot \mid x,a)\)
        }
        \\
        \bottomrule
   \end{tabular}
   }
\caption{
Examples of validation tasks expressed through responsiveness with safety applications.}
\label{Tab::RespTasks}
\end{table}

\subsection{Sampling Under Interactions}
\label{Sec::Algorithms}

Our framework falsifies, estimates, or tests the responsiveness of model outputs in three steps:
\begin{enumerate}[itemsep=0.2em]
    \item Specify an interaction model $\Imodel{\xp} = (\Aset{\xp}, \respdist{\xp,\cdot})$ 
    \item Generate a sample 
    \(\sampReach{\xp} := \{ \xq_{i} \}_{i=1}^n\) of feasible points under the interaction model $\Imodel{\xp}$
    \item Estimate responsiveness by querying the outputs of the model over the reachable set:
    \vspace{-.2em}
    \begin{align}
    \sampResp{\xp} := \tfrac{1}{n} \sum_{i=1}^n \mathbb{I}\left[ \clf{\xq_i} \in \ytarget{\xp} \right].
    \label{Eq::ResponsivenessEstimator}
    \end{align}
\end{enumerate}
This approach has a couple of practical benefits. It is model agnostic: it works for any model class and task because it estimates responsiveness by querying model outputs across $\sampReach{\xp}$. Moreover, we can amortize the cost of sampling by generating the sample $\sampReach{\xp}$ once and reusing it to estimate responsiveness across multiple models (e.g., to account for responsiveness during model selection).

\newcommand{\unsafe}[0]{\mathcal{X}_{\text{unsafe}}}

\newcommand{\block}[1]{S_{#1}}
\newcommand{\constSet}[1]{C_{{#1}}}
\newcommand{\sampleAction}[1]{\textsf{ProposeAction}({#1})}
\newcommand{\actLB}[2]{\text{LB}_{{#1}}({#2})}
\newcommand{\actUB}[2]{\text{UB}_{{#1}}({#2})}

\newcommand{\xlb}{\ell(\xp)}
\newcommand{\xub}{u(\xp)}
\algrenewcommand\algorithmiccomment[2][]{#1\hfill{\color{gray}\textit{\scriptsize{#2}}}}

\begin{algorithm}[t!]
\footnotesize
\begin{algorithmic}[1]\setstretch{1}
\footnotesize
\AlgNoIndent \textbf{Input} point $\xp$, interaction model $\Imodel{\xp} = (\Aset{\xp}, \respdist{\xp,\cdot})$, proposal distribution $q(\cdot \mid \xp)$, sample size $n$
\AlgNoIndent \textbf{Initialize} $\partition \gets \big\{ \{j\} \mid j \in \intrange{d} \big\}$, $\runningrset \gets \{\}$
\For{each constraint/effect $c$ in $\Imodel{\xp}$} \label{AlgStep::EnumConst}
    \State $\mathcal{M}_c \gets \{ \prt \in \partition \mid \prt \cap \scope{c} \neq \varnothing \}$ \Comment{get parts related to constraint/effect}
        \State $\partition \gets (\partition \setminus \mathcal{M}_c) \cup \big\{\bigcup_{\prt \in \mathcal{M}_c} S\big\}$
        \label{AlgStep::MergePart} \Comment{combine related parts}
\EndFor 
\While{$|\runningrset| < n$}
    \State $\xq \gets \xp$  \Comment{start from current point}
    \For{each part $\prt \in \partition$}
            \State Sample action part $\ab_\prt \sim q_\prt(\cdot \mid \xp_\prt)$ until $\rejector{\xp_\prt,\ab_\prt}$ passes
            \label{AlgStep::SampleAction} \Comment{sample feasible action in partition}
        \State $\bm{\varepsilon}_\prt \sim \respdist{S}(\xp_\prt,\ab_\prt)$ \label{AlgStep::SampleDownstream} \Comment{sample downstream effect within partition}
        \State $\xq_\prt \gets \xp_\prt + \ab_\prt + \bm{\varepsilon}_\prt$ \Comment{compute reachable sub-point}
    \EndFor
    \State $\runningrset \gets \runningrset \cup \{ \xq \}$ \label{AlgStep::AddPoint}
\EndWhile
\AlgNoIndent \textbf{Output} $\runningrset_n$, a sample of $n$ reachable points
\end{algorithmic}
\algrule{0.75\textwidth}{0.1pt}
$\rejector{\xp_\prt,\ab_\prt}$ solves a MIP to check if $\ab_\prt \in A_\prt(\xp_\prt)$ (\cref{Appendix::MIP}).
\vspace{-.5em}
\begin{equation*}
\min_{c} \; 0
\quad \text{s.t. } \bm{c} = \ab_\prt, \; \ab_\prt \in A_S(\xp_\prt)
\vspace{-0.5em}
\end{equation*}

\caption{Rejection Sampling for Reachable Points}
\label{Alg::ReachableSetSampling}
\vspace{-.25em}
\end{algorithm}

\paragraph{Sampling Algorithm}
The key challenge in designing a sampling-based approach is that the space of reachable points is highly irregular (see \cref{Fig::Overview}): it mixes discrete and continuous features, enforces joint constraints across features, and includes stochastic downstream effects. Standard samplers do not apply, and naive rejection sampling becomes inefficient as constraints accumulate~\citep{kroese2011handbook}.

We present an algorithm that samples reachable points efficiently by partitioning the feature space into conditionally independent parts (\cref{Alg::ReachableSetSampling}). Each part in $\partition := \{\prt_1, \ldots, \prt_k\}$ contains features coupled by a constraint or downstream effect, so that $\Aset{\xp} = \prod_{\prt \in \partition} A_{\prt}(\xp_\prt)$, where $A_{\prt}$ is the action set for part $\prt$. This strategy can dramatically improve the speed and scalability of rejection sampling in our setting, since each feasibility check involves only a small number of features.

For each part $\prt \in \partition$, we sample an action $\ab_\prt$ from the proposal distribution $q_\prt(\cdot \mid \xp_\prt)$ (\cref{AlgStep::SampleAction}) and pass it to $\rejector{\xp_\prt,\ab_\prt}$, which checks if $\ab_\prt$ obeys the actionability constraints within the part. If $\ab_\prt$ is feasible, we sample the downstream effect $\bm{\varepsilon}_\prt$ from $\respdist{S}(\xp_\prt,\ab_\prt)$. Once we have samples from all parts, we add $\xp + \ab + \bm{\varepsilon}$ to the reachable set $\runningrset$ (\cref{AlgStep::AddPoint}) and repeat until we have $n$ samples.

\section{Statistical Certification}
\label{Sec::InferenceGuarantees}

\todo[inline]{
- new title
- we have formal guarantees
- formal guarantees good because:
(i) interpretable findings/results
(ii) help practitioner design task (e.g., sample size)
}

A sample of reachable points $\sampReach{\xp}$ enable us to perform validation tasks with responsiveness in a wider range of use cases. However, it does not yield definitive conclusions about safety; it yields probabilistic claims.
Classical statistical estimation and testing procedures provide explicit interpretations of our probabilistic claims. These interpretations help practitioners make informed decisions when configuring the validation task (e.g., number of samples). Below, we describe the formal procedure for \emph{testing} and leave formal falsification and estimation procedures to \cref{Appendix::Guarantees}. 

\subsection{Hypothesis Testing}

We set up a hypothesis test to see if the responsiveness of a prediction exceeds or falls beneath a threshold value $\thresh > 0$. 
This is useful in settings where we may expect predictions to exhibit a minimum or maximum degree of responsiveness. 
For instance, in lending, we may want to detect preclusion---i.e., if individuals can change the features of a model to be approved for a loan---by testing if $\rsptrue{\xp} < \thresh$ for a small $\thresh$ (e.g., 0.01).

\begin{proposition}
    \label{Prop::Testing}
    Given a model $\clf{}: \X \to \Y$, we wish to test if its responsiveness at a point $\xp$ is below a threshold value $\thresh > 0$:
        \begin{align*}
            H_0:~\rsptrue{\xp} \geq \thresh \quad \text{vs.} \quad H_1:~\rsptrue{\xp} < \thresh
        \end{align*}
    Given responsiveness estimate $\rsphata{}$ constructed from a sample of $n$ reachable points $\sampReach{\xp}$ and a confidence level $\alpha \in (0, 1)$, we can claim that $\clf{\xp}$ is unresponsive whenever:
    \begin{align*}
        \rsphatmax{\xp}{\alpha}(n, \rsphata{}) := \betaQuant{1 - \alpha}{n \rsphata{} + 1}{n - n \rsphata{}} < \thresh,
    \end{align*}
    where $\betaQuant{\alpha}{a}{b}$ denotes the $\alpha^\textrm{th}$ quantile of a Beta distribution with parameters $a, b$.

    We can set the size of a reachable set to control the probability of correctly rejecting $H_0$---detecting unresponsive points. Assuming $\rsptrue{\xp} \leq \thresh - \altP$ for some margin $\altP \in (0, \thresh)$, we can ensure:
    \begin{align*}
        \prob{\text{Reject }H_0 \ | \ H_0 \text{ is False}} \geq 1 - \beta,
    \end{align*}
    by estimating responsiveness with a sample of at least $n^*$ reachable points where
    \begin{align*}
    n^* :=
    1 + \max \left\{
        n :
        \Pr{\rsphatmax{\xp}{\alpha}(n, \nicefrac{\nSucc}{n}) < \thresh } 
        < 1 - \beta
    \right\}, \text{ where } \nSucc \sim \mathsf{Binomial}(n, \thresh - \altP)
    \end{align*}
    
\end{proposition}

We can also deduce the minimum number of samples required for a valid test without making any assumptions on power $1-\beta$ and the margin $\delta$:
\begin{remark}
\label{Rem::MinSampSize}
    Consider testing if the responsiveness of $\clf{\xp}$ exceeds a threshold value $\thresh > 0$ via a hypothesis test in \cref{Prop::Testing}.
     Given the probability of incorrectly claiming $\clf{\xp}$ as unresponsive $\alpha > 0$, we need at least $n > \nicefrac{\log\alpha}{\log(1-\thresh)}$ reachable points to reject $H_0$.
\end{remark}

\subsection{Stress-Testing Interaction Models}

\todo[inline]{
-new title
when we have many interaction models we can compute
max rho(x[i]) over all interaction models 
min rho(x[i]) over all interaction models

these statistics are helpful for safety
min tells us the:
- when there is consensus -- useful for falsification (e.g., to show that we can find points that are responsive/unresponsive under any model)
- max tells us the breadth of the problem - useful to understand the downstream effects of disagreement (e.g., if everyone is affected by ambiguous responsiveness model then this means we gotta go back to the drawing board)
-more explicit math

algorithm:
-more comments
-rename partition in for loop to part
}

The interaction model $I(x)$ encodes the assumptions of practitioners explicitly. 
They must decide which features are actionable, what constraints govern joint changes, and whether there are downstream effects. 
These choices determine the set of feasible interactions and, consequently, responsiveness.
In practice, these choices are non-trivial, and it is difficult to gauge their impact on results a priori. 
Below, we present two post-hoc analyses of how these choices affect responsiveness.

\newcommand{\pval}[1]{p_{#1}}
\newcommand{\pvalAdj}[1]{\pval{#1}^\textrm{adj}}

\paragraph{Sensitivity Analysis}

In some cases, practitioners can disagree or be unsure about what the ``correct'' interaction model is.
In this case, we consider a family of interaction models $\intFamily$ and see if responsiveness is sensitive to changes in the interaction model. Given $\intFamily$, we can calculate:
\vspace{-.2em}
\begin{equation*}
    \rspub{\xp} = \max_{\Imodel{} \in \intFamily} \rho_I(\xp) \quad \text{or} \quad
    \rsplb{\xp} = \min_{\Imodel{} \in \intFamily} \rho_I(\xp) 
\end{equation*}
where $\rho_I(\xp)$ is the responsiveness of $\clf{\xp}$ under interaction model $\Imodel{\xp}$.
Suppose $\ytarget{\xp}$ is the set of safe outputs (e.g., prediction not gamed).
$\rspub{\xp}$ informs us if there is a \emph{consensus} of harm---i.e., $\clf{\xp}$ is unresponsive under all interaction models in $\intFamily$. This is useful when practitioners disagree on the interaction model. If we have a consensus, the disagreement is not a significant issue.
$\rsplb{\xp}$ informs us if there is a \emph{possibility} of harm---i.e., $\clf{\xp}$ is unresponsive under at least one  interaction model in $\intFamily$. This is useful when the correct interaction model is ambiguous. If there is a possibility of harm, practitioners must refine the exact interaction model and resolve ambiguities. Our sensitivity analysis for testing shares the same statistical certifications as \cref{Sec::InferenceGuarantees} (see \cref{Appendix::SensAnalysis}).

\paragraph{Identifying Drivers of (Un)responsiveness}

When a model output is unresponsive, we often want to know \emph{which constraints} in the interaction model are responsible.
For instance, one might suspect that a particular constraint (e.g., immutability of race) is preventing the output from changing, or may want to evaluate whether a new constraint (e.g., a policy rule) would materially affect responsiveness. 
We address this by systematically relaxing or removing individual constraints and measuring the resulting change in responsiveness.
More formally, given the original action set $\origAset{}$, we relax or remove the constraint of interest to obtain a relaxed action set $\relAset{} \supseteq \origAset{}$, and measure:
\(\Delta\rho := \rho_{\relAset{}}(\xp) - \rho_{\origAset{}}(\xp)\).
A large positive $\Delta\rho$ indicates that the ablated constraint was a key driver of unresponsiveness. Practitioners can repeat this across constraints to rank their contributions and identify actionable targets for model revision or policy changes. We detail the full procedure in \cref{Alg::ReachabilityProbe} in the Appendix and demonstrate it in \cref{Sec::Experiments}.

\subsection{Efficient Sampling with Side Information}

By default, we sample actions uniformly from $\Aset{\xp}$. This way, responsiveness ensures coverage across all feasible actions, rather than under specific assumptions about which actions are most likely. Coverage over actions is important in safety-critical applications because the actions that expose failures may not be known a priori; without reliable side information, we refrain from prematurely downweighting feasible actions that may later be consequential.
When additional information on informative regions of the reachable set is available, we can use \emph{importance sampling}: let $p(\cdot \mid \xp)$ denote the uniform density over $\Aset{\xp}$, and $q(\cdot \mid \xp)$ be a proposal distribution that concentrates probability on actions that practitioners believe are more informative. We reweight each sample to correct for the change in distribution:
\begin{align*}
    \hat{\rho}_q(\xp) := \tfrac{1}{n} \sum_{i=1}^{n} \tfrac{p(\ab_i \mid \xp)}{q(\ab_i \mid \xp)} \cdot \mathbb{I}\bigl[\clf{\xq_i} \in \ytarget{\xp}\bigr].
\end{align*}
This approach can accelerate the detection of failure modes in targeted settings by concentrating samples on actions that might be more likely to reveal safety violations. However, we do not use importance sampling in our framework by default because: 
(1) specifying a good proposal distribution $q$ is difficult, as it must be defined for each point $\xp$; and 
(2) $q$ must have support over the full action set (i.e., $q(\ab \mid \xp) > 0$ for all $\ab \in \Aset{\xp}$) since actions where $q$ has zero mass are never sampled and cannot be corrected by reweighting, biasing the estimator. 

\definecolor{detratecol}{HTML}{F5AF4D}
\definecolor{alarmcol}{HTML}{5495CF}
\definecolor{errcol}{HTML}{DB4743}
\definecolor{rgreen}{HTML}{7C873E}
\definecolor{rcream}{HTML}{FEF4D5}

Either issue can lead to unreliable validation results. Hence, we use uniform sampling as the default and view importance sampling as an extension for settings with reliable side information.

\section{Experiments and Demonstrations}
\label{Sec::Experiments}

We now demonstrate how our machinery can promote safety and reveal failure modes when developing or evaluating models using real-world data from salient applications. We include additional details in~\cref{Appendix::RecidivismPrediction,Appendix::ContentModeration,Appendix::LLMSafety}, an additional demonstration on testing responsiveness in organ transplant models in~\cref{Appendix::OrganTransplant}, and code \repourl{}.

\subsection{Convergence and Reliability}
\label{Sec::GermanDemo}
\begin{figure}[h!]
    \centering
    \providecommand{\ConvergenceData}{figures/german_convergence_data}
\providecommand{\FigFont}{\fontfamily{phv}\selectfont}
\providecolor{figblue}{HTML}{3B6FB6}
\providecolor{figred}{HTML}{E64B35}
\providecolor{figgreen}{HTML}{4C9F70}
\providecolor{figpurple}{HTML}{8C6BB1}
\providecolor{figamber}{HTML}{E1A730}
\providecolor{figambertext}{HTML}{C98F1A}  %
\providecolor{figink}{HTML}{1F2933}
\providecolor{figgrid}{HTML}{E3E7EB}
\tikzset{
  figticklabel/.style={font=\FigFont\fontsize{6}{7}\selectfont, color=figink!80, inner sep=0pt},
  figaxistitle/.style={font=\FigFont\fontsize{7}{8}\selectfont, color=figink, inner sep=0pt},
  figlabel/.style={font=\FigFont\fontsize{7}{8}\selectfont, color=figink, align=center, inner sep=0pt},
  figgridline/.style={color=figgrid, line width=0.5pt},
  figaxisline/.style={color=figink!70, line width=0.6pt},
  figcurve/.style={line width=0.9pt, line join=round},
}
\providecommand{\FigTextColor}[1]{\ifcsdef{\string\color@#1text}{#1text}{#1}}

\csdef{conv@minn}{29}  %
\csdef{conv@series}{detection,error,falsealarm}
\csdef{conv@tablen}{30}

\csdef{conv@detection@label}{Detection Rate}
\csdef{conv@detection@badge}{100,96}
\csdef{conv@detection@meaning}{Certifies an unresponsive prediction (\(\rho \leq \thresh - \delta\)) as unresponsive}
\csdef{conv@detection@value}{88.0\%}
\csdef{conv@detection@color}{figblue}
\csdef{conv@detection@line}{(10,0) (20,0) (30,88) (40,89) (50,92) (60,95) (70,93) (80,97) (90,95) (100,96)}
\csdef{conv@detection@band}{(10,0.0) (20,0.0) (30,93.7) (40,93.2) (50,94.7) (60,98.5) (70,95.7) (80,99.7) (90,98.5) (100,100.0) (100,91.8) (90,91.5) (80,94.3) (70,90.3) (60,91.5) (50,89.2) (40,84.8) (30,82.3) (20,0.0) (10,0.0)}

\csdef{conv@error@label}{Estimation Error}
\csdef{conv@error@badge}{100,11}
\csdef{conv@error@meaning}{Average gap between estimated and true responsiveness}
\csdef{conv@error@value}{4.4\%}
\csdef{conv@error@color}{figred}
\csdef{conv@error@line}{(10,7.57) (20,5.30) (30,4.39) (40,3.76) (50,3.33) (60,3.12) (70,2.82) (80,2.68) (90,2.52) (100,2.39)}
\csdef{conv@error@band}{(10,7.79) (20,5.44) (30,4.54) (40,3.88) (50,3.44) (60,3.20) (70,2.90) (80,2.76) (90,2.55) (100,2.48) (100,2.30) (90,2.47) (80,2.61) (70,2.76) (60,3.06) (50,3.22) (40,3.65) (30,4.24) (20,5.17) (10,7.35) (10,7.79)}

\csdef{conv@falsealarm@label}{False Alarm Rate}
\csdef{conv@falsealarm@badge}{100,0}
\csdef{conv@falsealarm@meaning}{Certifies a responsive prediction (\(\rho \geq \thresh\)) as unresponsive}
\csdef{conv@falsealarm@value}{0.1\%}
\csdef{conv@falsealarm@color}{figamber}
\csdef{conv@falsealarm@line}{(10,0.00) (20,0.00) (30,0.11) (40,0.11) (50,0.19) (60,0.06) (70,0.14) (80,0.12) (90,0.19) (100,0.12)}
\csdef{conv@falsealarm@band}{(10,0.00) (20,0.00) (30,0.23) (40,0.20) (50,0.37) (60,0.15) (70,0.27) (80,0.21) (90,0.32) (100,0.24) (100,0.01) (90,0.05) (80,0.03) (70,0.03) (60,0.00) (50,0.00) (40,0.00) (30,0.00) (20,0.00) (10,0.00)}

\def\convW{1.95in}   %
\def\convH{0.95in}   %
\def\convFloor{0}    %

\edef\convSeries{\csuse{conv@series}}
\tikzset{
  convbadge/.style={circle, draw=white, line width=0.4pt, text=white, inner sep=0pt,
    minimum size=7.5pt, font=\FigFont\fontsize{5.5}{5.5}\bfseries\selectfont},
}
\providecommand{\ConvBadge}[2]{\tikz[baseline=-0.6ex]\node[convbadge, fill=#1]{#2};}

\begin{minipage}[c]{0.44\linewidth}
\centering
\begin{tikzpicture}[x=\convW/90, y=\convH/(100-\convFloor), baseline=(current bounding box.center)]
\begin{scope}[xshift=-10*\convW/90]  %
  \foreach \t in {10,20,...,100} \draw[figgridline] (\t,\convFloor) -- (\t,100);
  \foreach \t in {0,25,50,75,100} \draw[figgridline] (10,\t) -- (100,\t);
  \foreach \t in {10,20,...,100} \node[figticklabel, anchor=north] at ([yshift=-5pt]\t,\convFloor) {\t};
  \foreach \t in {0,25,50,75,100} \node[figticklabel, anchor=east] at ([xshift=-3pt]10,\t) {\t\%};

  \draw[color=figink!55, line width=0.6pt, dash pattern=on 2pt off 1.5pt] (\csuse{conv@minn},\convFloor) -- (\csuse{conv@minn},100);
  \node[figlabel, anchor=west, text=figink!70] at ([xshift=2pt]\csuse{conv@minn},55) {\(n = \csuse{conv@minn}\)};

  \foreach \s in \convSeries {
    \edef\convclr{\csuse{conv@\s @color}}
    \edef\convband{\csuse{conv@\s @band}}
    \fill[fill=\convclr, fill opacity=0.18] plot coordinates {\convband} -- cycle;
  }
  \foreach \s in \convSeries {
    \edef\convclr{\csuse{conv@\s @color}}
    \edef\convline{\csuse{conv@\s @line}}
    \draw[figcurve, color=\convclr, mark=*, mark size=1.4pt,
      mark options={fill=\convclr, draw=white, line width=0.3pt}] plot coordinates {\convline};
  }
  \foreach \s [count=\k] in \convSeries {
    \edef\convclr{\csuse{conv@\s @color}}
    \node[convbadge, fill=\convclr] at ([xshift=8pt]\csuse{conv@\s @badge}) {\k};
  }

  \draw[figaxisline] (100,\convFloor) -- (10,\convFloor) -- (10,100);
  \node[figaxistitle, anchor=north] at ([yshift=-13pt]55,\convFloor) {Sample Size (\(n\))};
\end{scope}
\end{tikzpicture}%
\end{minipage}%
\hfill%
\begin{minipage}[c]{0.53\linewidth}
{\FigFont\fontsize{7}{8.5}\selectfont\color{figink}%
\renewcommand{\arraystretch}{1.3}%
\arrayrulecolor{figgrid}%
\begin{tabularx}{\linewidth}{@{}c@{\hspace{4pt}}>{\raggedright\arraybackslash}p{0.5in}@{\hspace{6pt}}>{\raggedright\arraybackslash\hyphenpenalty=10000}X@{\hspace{6pt}}r@{}}
  \multicolumn{2}{@{}l}{\fontsize{6}{7}\selectfont\color{figink!70}Metric}
    & \multicolumn{1}{@{}l@{\hspace{6pt}}}{\fontsize{6}{7}\selectfont\color{figink!70}What it measures}
    & {\fontsize{6}{7}\selectfont\color{figink!70}At \(n = \csuse{conv@tablen}\)} \\
  \midrule
  \ConvBadge{\csuse{conv@detection@color}}{1} & \csuse{conv@detection@label} & \csuse{conv@detection@meaning} & \csuse{conv@detection@value} \\
  \midrule
  \ConvBadge{\csuse{conv@error@color}}{2} & \csuse{conv@error@label} & \csuse{conv@error@meaning} & \csuse{conv@error@value} \\
  \midrule
  \ConvBadge{\csuse{conv@falsealarm@color}}{3} & \csuse{conv@falsealarm@label} & \csuse{conv@falsealarm@meaning} & \csuse{conv@falsealarm@value} \\
\end{tabularx}\arrayrulecolor{black}}%
\end{minipage}

    \caption{Estimation error and test reliability as we increase sample size to estimate responsiveness of XGBoost model predictions on the \textds{german} dataset~\citep{dua2019uci}. The vertical line marks $n=29$, the minimum number of samples required by \cref{Rem::MinSampSize} to certify unresponsiveness with $\alpha = 0.05, \thresh = 0.1, \delta = 0.05$. 
}
    \label{Fig::GermanDemoPlot}
\end{figure}

We start with a simple experiment to empirically demonstrate the reliability of our estimates and statistical guarantees in \cref{Sec::InferenceGuarantees}.
We work with the \textds{german} dataset~\citep{dua2019uci} to train an XGBoost classifier, and estimate the responsiveness of its predictions using $n=10,20,\ldots,100$ points (see \cref{Appendix::German} for the interaction model and detailed setup). Since the \textds{german} dataset is comprised of discrete features, we are able to compute the ground-truth value of responsiveness by enumeration. We compare our estimates to ground-truth values of $\rsptrue{\xp}$, showing how the estimation error, false alarm rate, and detection rate change as the sample size grows in \cref{Fig::GermanDemoPlot}.
\cref{Fig::GermanDemoPlot} shows that at $n=30$, the average estimation error is $4.4\%$, the false alarm rate is $0.1\%$, and the detection rate is $88.0\%$. These results show that the sample-size guarantee is conservative enough to keep false alarms rare while providing statistical power at modest sample sizes.

\subsection{Preclusion in Recidivism Prediction}
\begin{figure}[t!]
    \centering
    \providecommand{\AblationData}{figures/recidivism_ablation_data}
\input{figures/fig_style}
\input{\AblationData}

\edef\ablN{\csuse{abl@n}}
\def\ablW{1.7in}       %
\def\ablH{1.05in}       %
\def\ablGap{0.70in}     %

\begin{tikzpicture}[
  ablrowlabel/.style={figlabel, anchor=east},
  ablvalue/.style={figticklabel, anchor=west},
]

\begin{scope}[x=\ablW/100, y=\ablH/\ablN]
  \foreach \t in {20,40,...,100} \draw[figgridline] (\t,0) -- (\t,\ablN);
  \foreach \t in {0,20,...,100} \node[figticklabel, anchor=north] at ([yshift=-3pt]\t,0) {\t\%};
  \draw[figaxisline] (0,0) -- (0,\ablN);
  \foreach \i in {1,...,\ablN} {
    \pgfmathsetmacro\yc{\ablN-\i+0.5}
    \edef\ablclr{\csuse{abl@\i @color}}
    \fill[fill=\ablclr] (0,\yc-0.32) rectangle (\csuse{abl@\i @gain},\yc+0.32);
    \node[ablrowlabel] at ([xshift=-4pt]0,\yc) {\csuse{abl@\i @label}~(\csuse{abl@\i @constraint})};
    \node[ablvalue] at ([xshift=2.5pt]\csuse{abl@\i @gain},\yc) {\csuse{abl@\i @gain}\%};
  }
  \node[figaxistitle, anchor=north] at ([yshift=-11pt]50,0) {Change in Responsiveness};
\end{scope}

\begin{scope}[xshift=\ablW+\ablGap, x=\ablW/100, y=\ablH/100]
  \foreach \t in {20,40,...,100} {
    \draw[figgridline] (\t,0) -- (\t,100);
    \draw[figgridline] (0,\t) -- (100,\t);
  }
  \foreach \t in {0,20,...,100} \node[figticklabel, anchor=north] at ([yshift=-3pt]\t,0) {\t\%};
  \foreach \t in {20,40,...,100} \node[figticklabel, anchor=east] at ([xshift=-3pt]0,\t) {\t\%};
  \begin{scope}
    \clip (-1,-1) rectangle (101,101);
    \foreach \i in {\ablN,...,1} {
      \edef\ablclr{\csuse{abl@\i @color}}
      \edef\ablpts{\csuse{abl@\i @cdf}}
      \draw[figcurve, color=\ablclr] plot[const plot] coordinates {\ablpts};
    }
  \end{scope}
  \draw[figaxisline] (100,0) -- (0,0) -- (0,100);
  \foreach \i in {1,...,\ablN} {
    \ifcsdef{abl@\i @cdfnote}{%
      \edef\ablclr{\csuse{abl@\i @color}}%
      \edef\ablclr{\FigTextColor{\ablclr}}%
      \node[figlabel, text=\ablclr, anchor=\csuse{abl@\i @cdfnoteanchor}]
        at (\csuse{abl@\i @cdfnote}) {\csuse{abl@\i @cdfnotetext}};
    }{}
  }
  \node[figaxistitle, anchor=north] at ([yshift=-11pt]50,0) {Responsiveness};
  \node[figaxistitle, anchor=south, rotate=90] at ([xshift=-20pt]0,50) {Proportion of Individuals};
\end{scope}
\end{tikzpicture}
    \caption{Constraints driving preclusion in the recidivism prediction model.
    We run an ablation analysis of the interaction model: we relax one constraint at a time and measure how the responsiveness of individuals with fixed predictions ($n = 5{,}084$) changes.
    Left: mean change in responsiveness after relaxing each constraint; ``(I)'' marks an immutability constraint.
    Right: distribution of responsiveness after each relaxation.
    Relaxing the immutability of \texttt{age\_at\_release} and \texttt{drug\_abuser} largely resolves fixed predictions, indicating that the classifier relies on these demographic features rather than criminal history.}
    \label{Fig::RecidivismDemoResults}
\end{figure}

Many recidivism prediction models use features like \textfn{age} and \textfn{sex} to improve accuracy across subpopulations of defendants~\citep[see e.g.,][]{latessa2010creation,eaglin2017constructing,hester2019prior}. In practice, the features can lead models to predict recidivism \emph{by default}.  \citet{lawless2025understanding} show how a risk score developed by the Pennsylvania Sentencing Commission predicts that every male defendant under 21 will recidivate, regardless of their criminal history. This behavior can be seen as a failure in responsiveness where defendants who are predicted to recidivate cannot attain a different prediction even if they were to ``clear'' their criminal history. Failures like this rarely surface under existing practices. In this case, the risk score was a simple score that was designed to be scrutinized by a wide range of experts and that had undergone periodic audits by third parties~\citep{psc2017final}. Even so, the failure went undetected. 

In what follows, we show how our machinery would flag this failure and trace it to its root cause. We work with a dataset from the U.S. Department of Justice~\citep{icpsr1994} which contains labeled examples for $n = \num{29400}$ defendants from New York State. 
Here, each example consists of $d = 14$ features describing criminal history, demographics, sentence characteristics, and program participation. The label is set to $y_i = 1$ if defendant $i$ is rearrested within three years of release. We fit logistic regression and XGBoost models and test if they assign fixed predictions even when defendants ``clear'' their criminal history. 

Our interaction model permits \textfn{prior\_arrest} counts to decrease toward zero, program-participation features to increase, and treats demographic and sentence-related features as immutable. It also enforces internal consistency among criminal-history features: for example, setting $\textfn{prior\_arrests} = 5 \to 0$ requires $\textfn{n\_prior\_felony}$, as well as the other prior-history counts, to be set to zero. Given this model, we test for fixed predictions by checking whether $\rho(\xp) \leq \thresh$ with $\thresh = 0.01$, and $\alpha = 0.05$.

Our tests reveal that 17.3\% of defendants who are predicted to recidivate are assigned fixed predictions. \todo{Our test consists of checking if the predictions of a model change over a sample of valid interactions. Our test configuration supports the following formal claim for 2,258 individuals: say that each defendant who is predicted to recidivate would be able to clear their criminal history using at least $\thresh$\% of interactions. Then, there is at most a 5\% chance of observing these results. This guarantee reflects a default configuration that could be refined by increasing $\alpha$ and lowering $\thresh$.} %
One of the benefits of our approach is that we can readily understand the scope of failure by testing responsiveness across subpopulations or models. In this case, we find that the effect generalizes across training and test samples and arises when we work with simpler models such as logistic regression (since interactions are unchanged). We can pair such analysis with formal guarantees by repeating the sampling process or applying multiple testing correction. %

We use \cref{Alg::ReachabilityProbe} to trace the root causes of fixed predictions by relaxing individual constraints and measuring the resulting change in responsiveness. This analysis identifies effective points of intervention at both the population and individual level. As seen in \cref{Fig::RecidivismDemoResults}, relaxing immutability constraints on demographic and behavioral features (\texttt{age\_at\_release} and \texttt{drug\_abuser}) produces the largest gains in responsiveness on average, indicating that the model relies on these features rather than criminal history, and removing them would largely resolve the failure.

\subsection{Gaming in Content Moderation}

Modern content moderation systems often rely on machine learning models to limit misinformation or harassment at scale~\citep{gorwa2020algorithmic,gillespie2020content}.
In such settings, we may build models to predict if an account belongs to a  ``bot'' or ``human'' to guide follow-up actions~\citep[e.g., human review,][]{link2016human}.
Bot detection is difficult at scale: we need accurate models using many features, yet those features must be robust against bots that ``game'' the system to skirt detection. %

We consider a bot detection task from account characteristics in the \textds{twitterbot} dataset~\citep{gilani2017classification}. 
To build realistic attack models, we capture feasible interactions through non-separable constraints (e.g., forcing $\textfn{num\_tweets}$ to be 0 when $\textfn{inactivity}$ is 1).
We also assume some features (e.g., $\textfn{followers}$) are not actionable. 
Given the interaction model, we use an approach inspired by~\citep{zhang2015adversarial}: constructing classifiers that are robust to manipulation by penalizing or excluding certain features. We train a pool of penalized regression models over a large grid of $l_{1}$ and $l_{2}$ parameters using \texttt{glmnet}~\citep{friedman2010glmnet}. 

\begin{table*}[t!]
    \centering
    \small
    \resizebox{\linewidth}{!}{
\begin{tabular}{>{\emph}l l r r r r r}
\multicolumn{2}{c}{} &
\multicolumn{1}{c}{Models Considered} &
\multicolumn{2}{c}{Susceptibility to Gaming} &
\multicolumn{2}{c}{Best Model} \\

\cmidrule(l{3pt}r{3pt}){3-3}
\cmidrule(l{3pt}r{3pt}){4-5}
\cmidrule(l{3pt}r{3pt}){6-7}

\textbf{Procedure} &
\textbf{Description} &
\textbf{Robust / Total} &
\textbf{Reported} &
\textbf{Actual} &
\textbf{Test AUC} &
\textbf{$\Delta$} \\
\midrule

Baseline &
Ignore gaming &
901/901 &
63.0\% &
60.0\% &
0.763 &
--- \\

Manual &
Use only immutable features &
370/370 &
\textbf{0.0\%} &
\textbf{0.0\%} &
0.570 &
0.193 \\

Convex &
Cap gaming via convex relaxation &
687/901 &
\textbf{0.0\%} &
\textcolor{bad}{57.1\%} &
0.754 &
0.009 \\

Exact &
Cap gaming via exact model &
76/901 &
9.9\% &
9.9\% &
0.727 &
0.036 \\
\end{tabular}
}
    \caption{
Model selection on \textds{twitterbot} under four procedures. Each procedure selects the model with the highest validation AUC among those that pass its gaming check, and we report the test performance of that model. \emph{Baseline} returns an accurate model whose predictions are gameable. \emph{Manual} prevents gaming but loses accuracy. \emph{Convex} returns a model that appears robust but is gameable. Our method (\emph{Exact}) returns a robust model that loses little accuracy. We report the following measures: \emph{Robust / Total}, models that pass the check out of those considered; \emph{Reported}, the proportion of gameable predictions estimated by the procedure; \emph{Actual}, the proportion of gameable predictions under the exact interaction model; \emph{Test AUC}; and \emph{$\Delta$}, the drop in test AUC relative to \emph{Baseline}. All tests use $\thresh = 0.05$ and $\alpha = 0.05$.
}
\label{Table::TwitterResults}
\vspace{-1.5em}
\end{table*}

\cref{Table::TwitterResults} details three approaches to highlight practical challenges when building a well-performing model that also limits gaming. 
Although models trained under the \textit{Manual} procedure were all robust, they performed poorly with a Test AUC of 0.570. 
We also see that verifying responsiveness using a convex relaxation of the reachable set returns a well-performing model that \emph{appears} robust. 
In fact, the Test AUC of the model under the \textit{Convex} regime (0.754) exceeds that of the model chosen under \textit{Exact} (0.727).
However, we see that the \textit{Convex} procedure under-reports responsiveness: the \emph{perceived} proportion of responsive points is 0, when the proportion is actually $57.1\%$ under a faithful interaction model.

These results (1) show that there may exist a well-performing model that is robust to gaming without additional adversarial training and (2) highlight the importance of validating responsiveness with respect to accurate interactions. Procedures like \textit{Convex} can lead to unaccounted risk, where a model that is perceived as robust remains vulnerable to gaming.

\subsection{Evaluating Large Language Model Safety under Responsiveness}

We conclude with a demonstration of how responsiveness can be used in a broader range of settings and model classes. 
Here, we consider benchmarking in large language models (LLMs). 
Although models can be highly sensitive to changes in their inputs~\cite[see, e.g.,][]{lu2022fantastically}, standard practices in LLM evaluations are often static.
One particular downstream failure mode is selecting a model that might seem optimal in a standard static evaluation, but whose performance degrades under interventions. Responsiveness provides us with statistical tools to rigorously test model performance in this scenario.

\begin{table}[]
    \centering
    \resizebox{\linewidth}{!}{
    \begin{tabular}{lccccccccc}
     & & \multicolumn{2}{c}{$I_\mathrm{misspell}$} & \multicolumn{2}{c}{$I_\mathrm{roleplay}$} & \multicolumn{2}{c}{$I_\mathrm{chinese}$} & \multicolumn{2}{c}{$I_\mathrm{french}$} \\
    \cmidrule(l{3pt}r{3pt}){3-4} \cmidrule(l{3pt}r{3pt}){5-6} \cmidrule(l{3pt}r{3pt}){7-8} \cmidrule(l{3pt}r{3pt}){9-10}
    \textbf{Model} & \textbf{Compliance} & $\Delta$ & ($<$ 40.0\%) \textbf{p} & $\Delta$ & ($<$ 40.0\%) \textbf{p} & $\Delta$ & ($<$ 40.0\%) \textbf{p} & $\Delta$ & ($<$ 40.0\%) \textbf{p} \\
    \midrule
    OLMo 2 7B & 22.7\% & $+31.7$\,p.p. & \textcolor{bad}{0.998} & $-10.0$\,p.p. & $<\!0.001$ & $+31.7$\,p.p. & \textcolor{bad}{0.998} & $+33.0$\,p.p. & \textcolor{bad}{0.998} \\
    Llama 3.1 8B & 27.7\% & $-1.9$\,p.p. & $<\!0.001$ & $-0.6$\,p.p. & $<\!0.001$ & $+1.9$\,p.p. & 0.005 & $+1.9$\,p.p. & 0.005 \\
    Qwen 2.5 7B & 36.8\% & $+11.2$\,p.p. & \textcolor{bad}{1.000} & $-0.7$\,p.p. & $<\!0.001$ & $-1.6$\,p.p. & $<\!0.001$ & $+5.0$\,p.p. & \textcolor{bad}{1.000} \\
    Phi-3.5-mini & 42.4\% & $+20.1$\,p.p. & \textcolor{bad}{1.000} & $+6.3$\,p.p. & \textcolor{bad}{1.000} & $+19.0$\,p.p. & \textcolor{bad}{1.000} & $+6.1$\,p.p. & \textcolor{bad}{1.000} \\
        \end{tabular}
  }
  \vspace{.2em}
  \caption{Model selection under responsiveness on \textds{sorrybench}. We evaluate the rate of compliance with harmful prompts (lower is better) for several LLMs. For each interaction model, we evaluate responsiveness as compliance under interactions, e.g., misspelling, or translation to Chinese, and report the difference as $\Delta$. We conduct responsiveness tests at $\tau = 40\%$ and $\alpha = 0.05$, and report the p-values. We use Benjamini-Hochberg multiplicity correction within each LLM's set of tests.}
  \label{Tab::SorryBench}
  \vspace{-1.75em}
\end{table}

We consider a task of selecting a language model based on its compliance with harmful requests. To this end, we evaluate OLMo 2 7B~\cite{walsh2025olmo}, Llama 3.1 8B~\cite{grattafiori2024llama}, Qwen 2.5 7B~\cite{bai2023qwen}, and Phi-3.5 mini~\cite{haider2024phi}, in terms of their rate of compliance with harmful requests on the \textds{sorrybench} dataset~\cite{xie2025sorry}. We also measure their compliance rate under an interaction model which modifies the prompt to induce compliance. We consider several such interaction models: $I_\mathrm{misspell}$---the request is modified with misspellings, $I_\mathrm{roleplay}$---the request is framed as roleplay, $I_\mathrm{chinese}$ and $I_\mathrm{french}$---the request is translated from English to Chinese and French, respectively. For all interaction models, we consider the downstream effect $\varepsilon$ of the randomness in LLM sampling. The function $\clf{\xp}$ we evaluate is an indicator of whether a model complied with a request or not, as judged by the \textds{sorrybench} judge model, and $\ytarget{} = \{\texttt{comply}\}$. We evaluate \emph{average} responsiveness.
\vspace{-.2em}
\begin{equation*}
\E_{x \sim D}[\rho(\xp)] = \E_{\xp \sim D} \E_{\xq \sim X_I(\xp)}\big[\mathbb{I}[\clf{\xq} = \texttt{comply}]\big].
\end{equation*}
We find the base compliance rate of the OLMo model is the best, at 22.7\%. Then, we aim to test if, under interactions, the compliance rate is at most $\tau = 40\%$. First, we use the sample size calculation in \cref{Prop::Testing} to determine how many samples from the joint distribution of $(\xp, \xq)$, for $\xp \sim D, \xq \sim X_I(\xp)$, we need to test this at power $95\%$. E.g., we find that we need 79 samples for OLMo and 159 samples for Llama. For models with lower performance, we find that we need more than 2000 samples, so we focus on OLMo and Llama, which have the best base performance. Finally, we show the results of testing in \cref{Tab::SorryBench}. Even though OLMo has a lower base compliance rate on this benchmark and thus appears safer, its compliance under responsiveness increases substantially. In particular, we cannot ensure the compliance rate of OLMo is within the $\tau$ threshold.

\section{Concluding Remarks}
\label{Sec::ConcludingRemarks}

Machine learning is increasingly used in high-stakes systems with inputs that change through interactions from external actors. Yet our auditing practices and guarantees for model behavior under such interactions remain nascent ~\citep{nader1966unsafe}. Our work offers a concrete starting point for practitioners to reliably detect failures in how predictions respond to interactions.

Our approach requires practitioners to specify an interaction model based on domain knowledge, documentation, or policy rules. These specifications are assumptions that determine the constraints and downstream distributions. To guard against misspecified assumptions, we recommend repeating our tests over plausible interaction models. In practice, the outcome from this sensitivity analysis can help identify violations that are robust to assumptions surrounding actionability. Alternatively, it could also reveal that we fail to make any reliable claims over the plausible models.

\ifmlconfvenue{iclr}{%
\clearpage

\subsection*{AI Use Statement}
We used Claude \todo{models} for assistance writing and managing our codebase. All generated code was tested manually by the authors. We also used Claude \todo{models} for grammar checks and readability improvements. All language edits were manually reviewed by the authors.
}{}

\mlacknowledgments{%
This work was supported by funding from the National Science Foundation IIS 2040880, IIS 2313105, and the NIH Bridge2AI Center Grant U54HG012510.
}

\iftoggle{icml}{%
\clearpage
\section*{Impact Statement}
This paper presents work whose goal is to advance the field of machine learning. There are many potential societal consequences to this work, many of which we have discussed in the manuscript.

\bibliographystyle{ext/icml2026.bst}
}{
\mlbibliographystyle
}
{\small\bibliography{responsiveness_verification}}
{
}

\appendix
\crefalias{section}{appendix}
\crefalias{subsection}{subappendix}
\crefalias{subsubsection}{subsubappendix}
\appendixtoc

\clearpage 
\section{Additional Validation Procedures}
\label{Appendix::Guarantees}

\subsection{Statistical Certification for Estimation}
We begin with estimation, which supports decisions across the machine learning pipeline. Consider a lender or insurer who builds a model to approve loans or price premiums. They may fear that applicants can manipulate features to obtain favorable predictions without affecting true risk~\citep[][]{hardt2016strategic,hu2019strategic}. In such settings, model developers can estimate responsiveness to quantify the potential risks of manipulation and decide whether to remove manipulable features. By constructing a confidence interval around $\rsphat{\xp}$, they can bound the probability of making such decisions in error.

\begin{proposition}
    \label{Prop::Certify}
    Given a model $\clf{}: \X \to \Y$, a point $\xp$ and confidence level $\alpha \in (0,1)$, we wish to construct a confidence interval $\ci{}$ with:
    \begin{align*}
        \prob{\rsptrue{\xp} \in \ci_\alpha} \geq 1 - \alpha.
    \end{align*}
    We can construct a confidence interval that meets this condition based on an estimate $\rsphata{}$ from $n$ samples:
    \begin{align*}
        \ci_\alpha =%
        \bigl[\betaQuant{\tfrac{\alpha}{2}}{n\rsphata{}}{n - n\rsphata{} + 1}, \betaQuant{1 - \tfrac{\alpha}{2}}{n\rsphata{} + 1}{n - n\rsphata{}}\bigr]
    \end{align*}
    where $\betaQuant{\alpha}{a}{b}$ denotes the $\alpha^\textrm{th}$ quantile of a Beta distribution with parameters $a, b$.

    In this setting, we can restrict $\ci_\alpha = \ci_\alpha(n, \rsphata{})$ to a desired width $L$ by using a sample of at least $n^*$ reachable points where:
    \begin{align*}
        n^* := \min \left\{ n \in \mathbb{N}: \max_{s \in \intrange{n}} \left|\ci_\alpha\left(n, \frac{s}{n}\right)\right| \leq L \right \}.
    \end{align*}
\end{proposition}

The results in \cref{Prop::Certify} let us to gauge the number of samples practitioners need to achieve a desired level of precision. For a fixed level of risk $\alpha$, the sample size $n$ determines the level of uncertainty around the estimate $\rsphat{\xp}$---increasing $n$ decreases uncertainty: the estimate $\rsphat{\xp}$ is expected to converge to $\rsptrue{}$. This is because $\rsphat{\xp}$ is the empirical mean of i.i.d Bernoulli variables; consistency follows immediately from the Weak Law of Large Numbers.

For instance, if we know that we would like to estimate with a precision of given $\alpha = 0.05$ and desired width 0.1, we can calculate that $\nmin{0.05, 0.1} = 402$. To half the width at the same confidence level, we would need at least $\nmin{0.05, 0.05} = 1574$ samples.

\begin{example}

Consider a classification task where an online platform uses a model $\clf{}: \X \to \{0,1\}$ to detect bot accounts. To assess the potential cost of undetected bot accounts, the platform estimates responsiveness of bot predictions under $\clf{}$:
\begin{itemize}
    \item Target population $\{ \xp \in \X : \clf{\xp} = 1 \}$ 
    \item $\ytarget{\xp} = \{0\}$, classified as organic user
    \item $\Aset{\xp}$: actionability constraints
    \item $\alpha = 0.05$---95\% probability that $\ci{}$ contains true $\rsptrue{}$
    \item $L = 0.05$---$\ci{}$ can be at most 0.05 wide
\end{itemize}

Under these specifications, $\nmin{0.05, 0.05} = 1574$.

\end{example}

Falsification is equivalent to setting $\tau = 0$ for testing. If we observe one point in the sample that attains the target output, we can immediately falsify. When there are no samples that attain the target output, we can run testing with a very small $\tau$ (e.g., 0.001).

\subsection{Statistical Certification for Sensitivity Analysis}
\label{Appendix::SensAnalysis}
\paragraph{Constraint Analysis} Suppose practitioners disagree on the actionability constraints in the interaction model (e.g., in predicting loan amount, \textfn{num\_dependants} can change after a certain amount of time vs. a more condensed timeline assumes it cannot change).
In this case, we can test if interaction models in $\intFamily$ come to a \emph{consensus}. More formally, we consider the task results using: $$\rspub{\xp} = \max_{\Imodel{} \in \intFamily} \rho_{\Imodel{}}(\xp)$$
If the results hold for all $\Imodel{} \in \intFamily$, we can conclude that the conflict in specification is not significant.

\paragraph{Distribution Analysis} Suppose there is some ambiguity in the exact details of the interaction model (e.g., range of possible values for the parameters of the downstream distribution).
In this case, we can test if there exists a \emph{possibility} where we falsify or flag the prediction as unresponsive. More formally, we consider the task results using: $$\rsplb{\xp} = \min_{\Imodel{} \in \intFamily} \rho_{\Imodel{}}(\xp)$$
If there exists $\Imodel{} \in \intFamily$, where we falsify a prediction or claim that it is unresponsive, we can conclude that it is possible to have a safety violation under a realistic interaction model.

\paragraph{Hypothesis Testing} $\intFamily$ may not be countable (e.g., interaction models over an interval of downstream distribution parameters). We propose two ways to draw conclusions in testing depending on rejecting $H_0$ (i.e., sufficient evidence that $\rsptrue{\xp} < \thresh$) entails for safety:

\begin{plainlist}
    \item \emph{Flag}: There exists an interaction model $\Imodel{\xp} \in \intFamily$ where we reject $H_0$. More formally, we check if
    \begin{equation*}
        \min_{\Imodel{} \in \intFamily} \pvalAdj{\Imodel{}} < \alpha
    \end{equation*}
    where $\pvalAdj{\Imodel{}}$ is the $p$-value of the test under $\Imodel{}$ while controlling the family-wise error rate (e.g., using Bonferroni's or Holm's $p$-value adjustment~\citep{holm1979simple}. We would typically use this if $\ytarget{\xp}$ is a safe outcome (e.g., predicted to pay back loan), meaning rejecting $H_0$ is a safety violation.
\\
    \item \emph{Certify}: Check we reject $H_0$ for all interaction models $\Imodel{\xp} \in \intFamily$. More formally, we check if
    \begin{equation*}
        \max_{\Imodel{} \in \intFamily} \pval{\Imodel{}} < \alpha
    \end{equation*}
    where $\pval{\Imodel{}}$ is the $p$-value of the test under $\Imodel{}$. We would typically use this if $\ytarget{\xp}$ is an unsafe outcome (e.g., bots predicted as human users), meaning that rejecting $H_0$ is safe.
\end{plainlist}

Using the examples from the introduction, when testing for preclusion, we want to \emph{flag} fixed predictions, whereas in gaming, we want to \emph{certify} that the model is not susceptible to gaming across interaction models.

\clearpage 
\section{Description of the Mixed Integer Program}
\label{Appendix::MIP}

Our MIP checks if the sampled interaction $\ab'$ obeys the constraints in $\Aset{\xp}$. Although each $a_j$ for $j \in [d]$ obeys separable constraints: integrality, monotonicity and bounds, we must ensure that $\ab'$ does not violate joint constraints. Given $\xp$, a sampled interaction $\ab'$ and a set of constraints $\Aset{\xp}$, the routine solves a problem of form:
\begin{align}
\min_{\ab,\bm{c}} \; \|\ab\|_1 \quad \text{s.t.} \;\; \bm{c} = \ab',\; \bm{c} \in \Aset{\xp}
\label{Eq::CheckFeas} 
\end{align}
We implement \cref{Eq::CheckFeas} as a mixed-integer program that consists of a baseline formulation---enforcing separable constraints like bounds and monotonicity---and additional constraints to enforce joint conditions and deterministic downstream effects. The baseline formulation has the form:
\begin{samepage}
\begin{subequations}
\label{MIP::CF}
\footnotesize
\renewcommand{\arraystretch}{1.5}
\begin{equationarray}{@{}c@{}r@{\,}c@{\,}l>{\,}l>{\,}l@{\;}}
\min & \quad \displaystyle\sum_{j \in \intrange{d}} \left(\ajpos + \ajneg\right) \\[1.5em]
\st 
& c_j & = & a_j' & j \in \intrange{d}  & \mipwhat{match sampled aggregate interaction $\ab'$} \label{Con::CF::MatchInter}\\
& a_j, c_j & \in & [\interlb{j}{\xp}, \interub{j}{\xp}] & j \in \intrange{d} & \mipwhat{separable actionability bounds} \label{Con::CF::Bounds}\\
& \ajpos, \ajneg & \in & \R_{+} & j \in \intrange{d} & \mipwhat{positive, negative components of $a_j$} \label{Con::CF::Abs1}  \\
& a_j & = & \ajpos - \ajneg & j \in \intrange{d} & \mipwhat{absolute value reconstruction} \label{Con::CF::Recon}\\
& c_j & = & a_j & j \in \intrange{d} & \mipwhat{baseline aggregate change} \label{Con::CF::SetC}\\
& \sigma_j & \in & \B & j \in \intrange{d}  & \mipwhat{sign of $a_j$} \label{Con::CF::ASign} \\
& \ajpos & \geq & a_j & j \in \intrange{d} & \mipwhat{positive component of $a_j$} \label{Con::CF::AbsValue1}\\
& \ajneg & \geq & -a_j & j \in \intrange{d} & \mipwhat{negative component of $a_j$} \label{Con::CF::AbsValue2} \\
& \ajpos & \leq & \interub{j}{\xp} \sigma_j & j \in \intrange{d} & \mipwhat{only 1 of $\ajpos$ or $\ajneg$ can be positive} \label{Con::CF::SepUB}\\
& \ajneg & \leq & -\interlb{j}{\xp} (1-\sigma_j) & j \in \intrange{d} & \mipwhat{only 1 of $\ajpos$ or $\ajneg$ can be positive} \label{Con::CF::SepLB}\\
& \ab & \in & \Aset{\xp} & & \mipwhat{joint actionability constraints} \label{Con::CF::SepActionSet} 

\end{equationarray} 
\end{subequations}
\end{samepage}
where $\interlb{j}{\xp}, \interub{j}{\xp}$ denote the lower and upper bounds of actions for feature $j$ according to feature-level constraints.

The baseline formulation in \cref{MIP::CF} minimizes the $l_1$ norm of $\ab$, splitting $\ab$ into positive and negative parts $\ajpos, \ajneg \geq 0$ \eqref{Con::CF::Recon}, of which only one can be non-zero through \eqref{Con::CF::SepUB}--\eqref{Con::CF::SepLB}. This allows us to use this baseline formulation for both sampling and enumeration. Here, $\sigma_j$ is a boolean sign variable: $\sigma_j = 1$ allows positive movement and $\sigma_j = 0$ allows negative movement.

\eqref{Con::CF::MatchInter} stipulates that we intervene with sampled aggregate interaction $\ab'$---i.e., find an interaction $\ab$ whose aggregate change $\bm{c}$ satisfies the remaining constraints and is equal to $\ab'$. In the baseline model, \eqref{Con::CF::SetC} sets $\bm{c}=\ab$; deterministic downstream effects can replace this equality for linked target features. The remaining constraints enforce separable (constraint \eqref{Con::CF::Bounds}, \eqref{Con::CF::SepUB}, \eqref{Con::CF::SepLB}) and non-separable actionability constraints (constraint \eqref{Con::CF::SepActionSet}).

Below we provide all actionability constraints and their explicit formulations.

\subsection{Encoding Additive Linkage Constraints}

Additive linkage captures relationships in which an action on a source feature must equal (or upper/lower bound) a weighted sum of \emph{aggregate changes} in a set of target features. We use this construction to express conservation-like relationships between a source intervention and the total resulting change across one or more targets.

Given a source feature $k \in \intrange{d}$, a non-empty set of target features $T \subseteq \intrange{d} \setminus \{k\}$, a scalar source coefficient $\alpha_k \in \R \setminus \{0\}$, and target coefficients $\bm{\alpha}_T \in (\R \setminus \{0\})^{|T|}$, we add the following constraint to \cref{MIP::CF}:
\begin{align}
    \alpha_k a_k \;-\; \sum_{l \in T} \alpha_l c_l \;\square\; 0 \label{Eq::AddLink}
\end{align}
where $a_k$ is the action on the source, $c_l$ is the aggregate change in target $l$ (as defined in \cref{Eq::AggChange}), and $\square \in \{=, \geq, \leq\}$ specifies whether the linkage is enforced as equality or inequality. Note that the use of $c_l$ allows additive linkage to compose with directional linkage on the same target features.

\subsection{Encoding Directional Linkage Constraints} We often encounter features where intervening on them has a direct (and sometimes deterministic) effect on other features. For example, in \cref{Table::ActionSetTwitter}, joint constraint 4 stipulates that \textfn{urls\_count} increases at most as the change in \textfn{num\_tweets}. Here, the ``source variable''---the source of the effect---is \textfn{num\_tweets} and the ``target variable''---the feature affected---is \textfn{urls\_count}.

We capture this effect, called \emph{Directional Linkage}, by adding additional constraints to \cref{MIP::CF}. Given source feature $k \in \intrange{d}$, a non-empty set of target features $T \subseteq \intrange{d} \setminus \{k\}$, and a scale vector $\tilde{\bm{s}} \in (\R \setminus \{0\})^{|T|+1}$ with one entry per feature in $\{k\} \cup T$, we define the effective per-target scales $s_l := \tilde{s}_l / \tilde{s}_k$ for each $l \in T$. We then introduce one auxiliary variable $b_l$ per target capturing the induced change, and add the following constraints to \cref{MIP::CF}:
\begin{align}
    b_{l} - s_l \cdot a_k &= 0 \label{Eq::DirectLink} \\
    c_l - a_l - b_{l} &= 0 \label{Eq::AggChange}
\end{align}
for each target feature $l \in T$, where $b_l$ is the change in feature $l$ induced by the source action $a_k$, and $c_l$ represents the aggregate change in $l$.

The auxiliary variable $b_l$ is bounded by the propagation of the source action bounds $[\, -\interlb{k}{\xp},\, \interub{k}{\xp}\,]$ through the scale $s_l$, namely
\begin{align}
    \min\!\big(s_l \interub{k}{\xp},\, -s_l \interlb{k}{\xp}\big) \;\leq\; b_l \;\leq\; \max\!\big(s_l \interub{k}{\xp},\, -s_l \interlb{k}{\xp}\big), \label{Eq::DirectLink::Bounds}
\end{align}
which handles both signs of $s_l$. By default, the linkage also \emph{widens} the aggregate-change bounds on $c_l$ by the corresponding $b_l$ bounds, so that $c_l$ can absorb the induced effect on top of any direct action $a_l$. When the linkage is configured to preserve the original target bounds, this widening step is skipped and the original $c_l$ bounds are kept.

We can also substitute the equality in \cref{Eq::DirectLink} with inequalities. The aforementioned example with \textfn{num\_tweets} and \textfn{urls\_count} is a case where the relationship is an inequality ($\leq$) and $s_l = 1$.

\subsection{Encoding If-Then Constraints}

If-then constraints enforce conditional relationships of the form ``if the action on feature $k$ pushes $\xp_k$ above a threshold, then the post-action value of feature $l$ must satisfy a consequent condition''. This is useful for situations such as requiring a minimum balance level once a checking account is opened, or requiring a minimum program completion level once enrollment changes.

We capture an antecedent action threshold $v^\textrm{if}$ on feature $k$, with the antecedent active when $a_k \geq v^\textrm{if}$, and a consequent on the \emph{post-action feature value} $x_l + a_l$ given as either $x_l + a_l \geq v^\textrm{then}$ or $x_l + a_l = v^\textrm{then}$. We introduce a binary indicator $u \in \B$ that activates when the antecedent holds, and add the following big-$M$ constraints to \cref{MIP::CF}:
\begin{subequations}
\label{MIP::IfThen}
\begin{align}
    M^\textrm{fwd} u - a_k &\geq -v^\textrm{if} + \epsilon \label{Eq::IfThen::Fwd}\\
    -M^\textrm{bwd} u + a_k &\geq v^\textrm{if} - M^\textrm{bwd} \label{Eq::IfThen::Bwd}\\
    a_l - M^\textrm{lb}_l u &\geq \tilde{v}^\textrm{then} - M^\textrm{lb}_l \label{Eq::IfThen::ThenLB}
\end{align}
\end{subequations}
where $\tilde{v}^\textrm{then} = v^\textrm{then} - x_l$ rewrites the consequent feature value as a required action magnitude, $\epsilon > 0$ is a small tolerance, and the big-$M$ constants are tightened from the bounding box on $\ab$:
\begin{align*}
    M^\textrm{fwd} = \interub{k}{\xp} - v^\textrm{if} + \epsilon, \quad
    M^\textrm{bwd} = v^\textrm{if} + \interlb{k}{\xp}, \quad
    M^\textrm{lb}_l = \tilde{v}^\textrm{then} + \interlb{l}{\xp}.
\end{align*}
\eqref{Eq::IfThen::Fwd} and \eqref{Eq::IfThen::Bwd} together force $u = \indic{a_k \geq v^\textrm{if}}$, while \eqref{Eq::IfThen::ThenLB} enforces the consequent lower bound on $a_l$ (equivalent to the lower bound on $x_l + a_l$) only when $u = 1$. Note that our implementation always uses this action-threshold trigger for the antecedent, regardless of whether the antecedent condition is specified with equality or inequality sense. When the consequent uses an equality sense, we additionally add an upper bound:
\begin{align}
    a_l + M^\textrm{ub}_l u &\leq \tilde{v}^\textrm{then} + M^\textrm{ub}_l \label{Eq::IfThen::ThenUB}
\end{align}
with $M^\textrm{ub}_l = \interub{l}{\xp} - \tilde{v}^\textrm{then}$.

\subsection{Encoding One-Hot Constraints}

One-hot constraints maintain the encoding of a categorical attribute that has been split into a collection of binary dummies. Given a categorical attribute $Z$ with $m$ levels encoded as boolean features $x_{j_1}, \ldots, x_{j_m}$ where $x_{j_k} = \indic{Z = z_k}$, valid feature vectors must satisfy a cardinality requirement on the dummies.

Given the index set $J \subseteq \intrange{d}$ of dummies and a cardinality limit $L \in \intrange{|J|}$, we add a single linear constraint to \cref{MIP::CF}:
\begin{align}
    \sum_{j \in J} a_j \;\square\; L - \sum_{j \in J} x_j \label{Eq::OneHot}
\end{align}
where $\square$ is $=$ when the encoding is \emph{exact} (exactly $L$ dummies must be active) and $\leq$ when the encoding is \emph{at most $L$}. The right-hand side rewrites the cardinality requirement on $\xp + \ab$ in terms of the action $\ab$, so that any feasible action preserves the encoding.

\subsection{Encoding Ordinal Constraints}

Ordinal constraints maintain the encoding of an ordinal attribute with $m$ levels that has been one-hot encoded. They differ from one-hot constraints in that we may also wish to restrict transitions between levels (e.g., an ordinal attribute representing tenure can only increase).

Ordinal constraints are a special case of the reachability constraint encoding instantiated as follows. The set of feasible values $V$ contains every length-$m$ binary vector with exactly one entry equal to one. We construct the reachability matrix $E \in \{0,1\}^{|V| \times |V|}$ from a step direction $\sigma \in \{-1, 0, 1\}$:
\begin{align*}
    E_{ij} = \begin{cases}
        \indic{i \leq j} & \text{if } \sigma = +1 \text{ (level can only increase)} \\
        \indic{j \leq i} & \text{if } \sigma = -1 \text{ (level can only decrease)} \\
        1 & \text{if } \sigma = 0 \text{ (free)}
    \end{cases}
\end{align*}
where $V$ is ordered by ascending level. We then add \cref{Eq::Reach1,Eq::ReachUB,Eq::Reach2} from the reachability constraint subsection below using these $V$ and $E$.

\subsection{Encoding Reachability Constraints}

Reachability constraints generalize one-hot, ordinal and thermometer encodings to arbitrary subsets of features that are restricted to a finite set of jointly feasible values, with optional restrictions on which values can be reached from a given starting point. Many encoding constraints in our framework reduce to a reachability constraint with an appropriate value set $V$ and reachability matrix $E$.

Given an ordered set of feasible values $V \subseteq \R^{|J|}$ over a feature subset $J \subseteq \intrange{d}$ and a reachability matrix $E \in \{0,1\}^{|V| \times |V|}$ with $E_{ij} = 1$ iff $v_j$ is reachable from $v_i$, we add the following constraints to \cref{MIP::CF}. Let $i^* \in \intrange{|V|}$ denote the index of the value matching $\xp_J$ (i.e., $v_{i^*} = \xp_J$), and let $e_k = E_{i^*, k}$ be the reachability indicator for value $v_k$ from the current point. We introduce indicators $u_k \in \B$ for $k \in \intrange{|V|}$ and require:
\begin{align}
    \sum_{k \in \intrange{|V|}} u_k &= 1 \label{Eq::Reach1} \\
    u_k &\leq e_k && \forall\, k \in \intrange{|V|} \label{Eq::ReachUB}\\
    a_j &= \sum_{k \in \intrange{|V|}} e_k \, (v_{k,j} - x_j) \, u_k && \forall\, j \in J \label{Eq::Reach2}
\end{align}
\eqref{Eq::Reach1} ensures that exactly one feasible value is selected, \eqref{Eq::ReachUB} forbids selecting unreachable values---without it, the model could pick $u_k = 1$ for some $v_k$ with $e_k = 0$, which would zero out the right-hand side of \eqref{Eq::Reach2} and falsely report $\ab_J = 0$ while ``selecting'' an unreachable destination---and \eqref{Eq::Reach2} pins each component of $\ab$ on $J$ to the action that moves $\xp_J$ to the selected value $v_k$. In our CPLEX implementation, \eqref{Eq::ReachUB} is enforced directly by setting the upper bound of each indicator $u_k$ to $e_k$.

\subsection{Encoding Switch Constraints}

Switch constraints capture mutability that is conditional on a binary ``switch'' feature. Concretely, when the switch is on, a designated set of target features cannot change; when the switch is off, those targets may (and optionally must) change. For example, in the \textds{german} dataset, \textfn{CheckingAcct$\geq$0} should only be permitted to vary when \textfn{CheckingAcct\_exists} is \texttt{True}.

By default, our implementation treats the switch state at the input point $\xp$ as a precondition: the constraint requires that $\xp$ already satisfies the implied target invariant (targets are zero when the switch is on at $\xp$, and---if force-change is enabled---non-zero when the switch is off at $\xp$). The MIP encoding below is added only when this precondition is met; a permissive mode that skips this check is also supported.

Given a boolean switch feature $k \in \intrange{d}$, an ``on value'' $\nu \in \{0,1\}$, a set of target features $T \subseteq \intrange{d} \setminus \{k\}$, and per-target step bounds $A^+_l, A^-_l \geq 0$ derived from $\interub{l}{\xp}, \interlb{l}{\xp}$, we introduce a binary indicator $w \in \B$ that tracks the \emph{post-action} switch state. We then add the following constraints to \cref{MIP::CF}:
\begin{subequations}
\label{MIP::Switch}
\begin{align}
    w &= \begin{cases}
        x_k + a_k & \text{if } \nu = 1 \\
        1 - x_k - a_k & \text{if } \nu = 0
    \end{cases} \label{Eq::Switch::Set}\\
    a^+_l + A^+_l \, w &\leq A^+_l && \forall\, l \in T \label{Eq::Switch::Pos}\\
    a^-_l + A^-_l \, w &\leq A^-_l && \forall\, l \in T \label{Eq::Switch::Neg}
\end{align}
\end{subequations}
where $a^+_l, a^-_l$ are the positive and negative parts of $a_l$ from \eqref{Con::CF::Recon}. \eqref{Eq::Switch::Set} forces $w = 1$ exactly when the post-action switch state $x_k + a_k$ matches the on-value $\nu$, and \eqref{Eq::Switch::Pos}--\eqref{Eq::Switch::Neg} together fix $a_l = 0$ for every target $l \in T$ whenever $w = 1$.

When we additionally require that targets must change when the switch is off (i.e., $w = 0$), we add a single forcing constraint:
\begin{align}
    \sum_{l \in T} \big(a^+_l + a^-_l\big) + \delta \, w &\geq \delta \label{Eq::Switch::Force}
\end{align}
where $\delta$ is slightly less than the minimum step size across targets in $T$. When $w = 1$, this constraint is slack; when $w = 0$, it forces the total absolute movement on the targets to be at least $\delta$.

\subsection{Encoding Thermometer-Encoding Constraints}

Datasets often include features that are based on thresholds. These features are often encoded like unary codes, a number of ones followed by zeros. For example, in \cref{Table::ActionSetTwitter}, \textfn{age\_of\_account\_geq} has a thermometer encoding with thresholds at 180, 365, 730 and 1825 days. Hence there are five possible encoding values:
\begin{enumerate}
    \item $[0, 0, 0, 0]$: account is less than 180 days old
    \item $[1, 0, 0, 0]$: account is older than 180 days but less than 365 days old
    \item $[1, 1, 0, 0]$: account is older than 365 days but less than 730 days old
    \item $[1, 1, 1, 0]$: account is older than 730 days but less than 1825 days old
    \item $[1, 1, 1, 1]$: account is more than 1825 days old
\end{enumerate}

Given an ordered set of feasible values $V$ as above, we define a reachability matrix $E \in \{0, 1\}^{|V| \times |V|}$, where the $(i,j)$-th entry of $E$ is 1 when we can reach the $j$-th element of $V$ from its $i$-th element and 0 otherwise. There are three possibilities for $E$: an upper triangular matrix of ones (when the underlying value can only increase), a lower triangular matrix of ones (when it can only decrease), or an all-ones matrix (when the value can move freely). For example, \textfn{age\_of\_account\_geq} additionally has a monotonicity constraint---age can only increase---so $E$ is an upper triangular matrix of ones (i.e., one can reach $[1,0,0,0]$ from $[0,0,0,0]$, but not the other way around).

A thermometer-encoded feature subset is then encoded in the MIP exactly as a reachability constraint: we add \cref{Eq::Reach1,Eq::ReachUB,Eq::Reach2} from the reachability constraint subsection above with the value set $V$ and reachability matrix $E$ defined here. As with reachability, the upper bound $u_k \leq e_k$ in \eqref{Eq::ReachUB} is enforced directly via the indicator's variable bound.

\clearpage
\section{Supplementary Material for Section \ref{Sec::Algorithms}}
\label{Appendix::Algorithms}
\label{Appendix::Proofs}
\label{Appendix::OmittedProofs}

\subsection{Algorithm for Constraint Relaxation}

\begin{algorithm}[h!]
\caption{Paired Reachability Probe}
\label{Alg::ReachabilityProbe}
\begin{algorithmic}[1]\footnotesize
\Require{
Point $\xp$,
sample size $n$,
original action set $\origAset{\xp}$,
relaxed action set $\relAset{\xp}$,
classifier $\clf{}$
}
\State $y_0 \gets \clf{\xp}$
\State $\mathcal{P} \gets \text{Sample }(n \text{ proposals from } \relAset{\xp} \text{ around } \xp)$
\State $\mathsf{feas} \gets \mathbf{1}_n$
\State $\mathsf{feas}^{\textrm{rel}} \gets \mathbf{1}_n$
\For{each partition $p$ in $\origAset{\xp}$}
    \State $\mathsf{feas} \gets \mathsf{feas} \land \text{IsFeasible}(p, \origAset{\xp}, \mathcal{P}[p])$
    \State $\mathsf{feas}^{\textrm{rel}} \gets \mathsf{feas}^{\textrm{rel}} \land \text{IsFeasible}(p, \relAset{\xp}, \mathcal{P}[p])$
\EndFor
\State $\hat{y} \gets -\mathbf{1}_n$, \quad $\hat{y}^{\textrm{rel}} \gets -\mathbf{1}_n$
\State $\hat{y}[\mathsf{feas}] \gets \clf{\mathcal{P}[\mathsf{feas}]}$
\State $\hat{y}^{\textrm{rel}}[\mathsf{feas}^{\textrm{rel}}] \gets \clf{\mathcal{P}[\mathsf{feas}^{\textrm{rel}}]}$
\State \Return $\big(y_0,\, \mathsf{feas},\, \mathsf{feas}^{\textrm{rel}},\, \hat{y},\, \hat{y}^{\textrm{rel}}\big)$
\end{algorithmic}
\end{algorithm}

\clearpage
\section{Supporting Material for the \textds{german} Study}
\label{Appendix::German}

In \cref{Fig::GermanDemoPlot}, we conduct a study on sample size $n$ to (1) validate our responsiveness estimation and testing procedure outlined in \cref{Sec::Algorithms}, and (2) determine their reliability under various sample sizes. We work with a discrete dataset, \textds{german}, where we can fully enumerate reachable sets using the enumeration procedure from \citet{kothari2024prediction}. The enumerated reachable sets provide ground truth responsiveness proportions. We compare our results to determine the error of our \emph{estimation} procedure and the precision (in the main body, we refer to it as ``reliability'' for simplicity) of our \emph{testing} procedure under XGBoost (\XGB{}).

The \textds{german} dataset is a credit dataset originally compiled in 1994 that is publicly available through the UCI Machine Learning Repository \citep{dua2019uci}. It contains $n=\num{1000}$ de-identified instances, each representing a credit applicant. It includes $d=20$ categorical or discrete features, providing insights into aspects such as loan history, demographic information (including gender, age, and marital status), occupation, and past payment behavior. The objective is to predict whether an applicant is a ``good'' $(y_i=1)$ or ``bad'' $(y_i=0)$ credit customer. We note that the dataset does not have missing values, and have adapted some feature names for clarity.

\subsection{Interaction Model}

We consider an interaction model where each applicant can intervene on current features like account balances, but not history nor credit related features. For example, \textfn{Housing=Owner} is not actionable since one cannot go from renting to buying without additional loans. This interaction model is conservative and is intended to capture indisputable actionability constraints.
In total, our dataset contains 36 features of which 9 are actionable and 10 are mutable. There a total of four constraints: two Directional Linkage, and two Thermometer Encoding constraints:
\begin{bulletlist}
    \item Directional Linkage constraints in this interaction model govern downstream effects on \textfn{Age} from 1) \textfn{YearsAtResidence} and 2) \textfn{YearsEmployed$\geq$1}, which form a partition.
    \item Thermometer Encoding constraints enforce conceptual requirements in this dataset - 1) requiring \textfn{CheckingAcct$\geq$0}$=$\texttt{True} to be reachable only if \textfn{CheckingAcct\_exists} is also \texttt{True}, and 2) requiring \textfn{SavingsAcct$\geq$100}$=$\texttt{True} to be reachable only if \textfn{SavingsAcct\_exists} is \texttt{True}.
\end{bulletlist}

These lead to 31 partitions.

We present a list of all features and their corresponding feature-level constraints in \cref{Table::ActionSetGerman} and list the non-separable joint constraints below it.

\begin{table}[h]
\centering
\resizebox{0.85\linewidth}{!}{
\begin{tabular}{lllllllr}
\textheader{Name} & \textheader{Type} & \textheader{LB} & \textheader{UB} & \textheader{Actionable} & \textheader{Sign} & \textheader{Joint Constraints} & \textheader{Partition ID} \\
\midrule
\rowcolor{gray!30} \textfn{Age} & $\mathbb{Z}$ & 19 & 75 & No &  & 1, 2 & 0 \\
\textfn{YearsAtResidence} & $\mathbb{Z}$ & 0 & 7 & Yes & $+$ & 1 & 0 \\
\textfn{YearsEmployed$\geq$1} & $\{0,1\}$ & 0 & 1 & Yes & $+$ & 2 & 0 \\
\textfn{CheckingAcct\_exists} & $\{0,1\}$ & 0 & 1 & Yes & $+$ & 3 & 30 \\
\textfn{CheckingAcct$\geq$0} & $\{0,1\}$ & 0 & 1 & Yes & $+$ & 3 & 30 \\
\textfn{SavingsAcct\_exists} & $\{0,1\}$ & 0 & 1 & Yes & $+$ & 4 & 31 \\
\textfn{SavingsAcct$\geq$100} & $\{0,1\}$ & 0 & 1 & Yes & $+$ & 4 & 31 \\
\rowcolor{gray!30} \textfn{Male} & $\{0,1\}$ & 0 & 1 & No &  & -- & 1 \\
\rowcolor{gray!30} \textfn{Single} & $\{0,1\}$ & 0 & 1 & No &  & -- & 2 \\
\rowcolor{gray!30} \textfn{ForeignWorker} & $\{0,1\}$ & 0 & 1 & No &  & -- & 3 \\
\rowcolor{gray!30} \textfn{LiablePersons} & $\mathbb{Z}$ & 1 & 2 & No &  & -- & 4 \\
\rowcolor{gray!30} \textfn{Housing$=$Renter} & $\{0,1\}$ & 0 & 1 & No &  & -- & 5 \\
\rowcolor{gray!30} \textfn{Housing$=$Owner} & $\{0,1\}$ & 0 & 1 & No &  & -- & 6 \\
\rowcolor{gray!30} \textfn{Housing$=$Free} & $\{0,1\}$ & 0 & 1 & No &  & -- & 7 \\
\rowcolor{gray!30} \textfn{Job$=$Unskilled} & $\{0,1\}$ & 0 & 1 & No &  & -- & 8 \\
\rowcolor{gray!30} \textfn{Job$=$Skilled} & $\{0,1\}$ & 0 & 1 & No &  & -- & 9 \\
\rowcolor{gray!30} \textfn{Job$=$Management} & $\{0,1\}$ & 0 & 1 & No &  & -- & 10 \\
\rowcolor{gray!30} \textfn{CreditAmt$\geq$1000K} & $\{0,1\}$ & 0 & 1 & No &  & -- & 11 \\
\rowcolor{gray!30} \textfn{CreditAmt$\geq$2000K} & $\{0,1\}$ & 0 & 1 & No &  & -- & 12 \\
\rowcolor{gray!30} \textfn{CreditAmt$\geq$5000K} & $\{0,1\}$ & 0 & 1 & No &  & -- & 13 \\
\rowcolor{gray!30} \textfn{CreditAmt$\geq$10000K} & $\{0,1\}$ & 0 & 1 & No &  & -- & 14 \\
\rowcolor{gray!30} \textfn{LoanDuration$\leq$6} & $\{0,1\}$ & 0 & 1 & No &  & -- & 15 \\
\rowcolor{gray!30} \textfn{LoanDuration$\geq$12} & $\{0,1\}$ & 0 & 1 & No &  & -- & 16 \\
\rowcolor{gray!30} \textfn{LoanDuration$\geq$24} & $\{0,1\}$ & 0 & 1 & No &  & -- & 17 \\
\rowcolor{gray!30} \textfn{LoanDuration$\geq$36} & $\{0,1\}$ & 0 & 1 & No &  & -- & 18 \\
\rowcolor{gray!30} \textfn{LoanRate} & $\mathbb{Z}$ & 1 & 4 & No &  & -- & 19 \\
\textfn{HasGuarantor} & $\{0,1\}$ & 0 & 1 & Yes & $+$ & -- & 20 \\
\rowcolor{gray!30} \textfn{LoanRequiredForBusiness} & $\{0,1\}$ & 0 & 1 & No &  & -- & 21 \\
\rowcolor{gray!30} \textfn{LoanRequiredForEducation} & $\{0,1\}$ & 0 & 1 & No &  & -- & 22 \\
\rowcolor{gray!30} \textfn{LoanRequiredForCar} & $\{0,1\}$ & 0 & 1 & No &  & -- & 23 \\
\rowcolor{gray!30} \textfn{LoanRequiredForHome} & $\{0,1\}$ & 0 & 1 & No &  & -- & 24 \\
\rowcolor{gray!30} \textfn{NoCreditHistory} & $\{0,1\}$ & 0 & 1 & No &  & -- & 25 \\
\rowcolor{gray!30} \textfn{HistoryOfLatePayments} & $\{0,1\}$ & 0 & 1 & No &  & -- & 26 \\
\rowcolor{gray!30} \textfn{HistoryOfDelinquency} & $\{0,1\}$ & 0 & 1 & No &  & -- & 27 \\
\textfn{HistoryOfBankInstallments} & $\{0,1\}$ & 0 & 1 & Yes & $+$ & -- & 28 \\
\textfn{HistoryOfStoreInstallments} & $\{0,1\}$ & 0 & 1 & Yes & $+$ & -- & 29 \\
\end{tabular}

}
\caption{
Interaction Model for the processed \textds{german} dataset. 
\textbf{Type} indicates the feature type ($\Z$ for integer, $\{0,1\}$ for binary). 
\textbf{LB}, \textbf{UB} are the lower and upper bounds for the feature.
\textbf{Actionable} indicates whether the feature is globally actionable. Non-actionable features are highlighted.
\textbf{Sign} indicates monotonicity constraints---whether feature can only increase (+) or decrease (-).
\textbf{Joint Constraints} are a list non-separable constraint indices (listed below table) it is tied to (if any).
\textbf{Partition ID} indicates which partition the feature belongs to.
}
\label{Table::ActionSetGerman}
\end{table}

\begin{constraints}
\item DirectionalLinkage: Actions on \textfn{YearsAtResidence} will induce  actions on [`\textfn{Age}']. Each unit change in \textfn{YearsAtResidence} leads to a unit change in \textfn{Age}
\item DirectionalLinkage: Actions on \textfn{YearsEmployed$\geq$1} will induce  actions on [`\textfn{Age}']. Each unit change in \textfn{YearsEmployed$\geq$1} leads to a unit change in \textfn{Age}
\item ThermometerEncoding: Actions on [\textfn{CheckingAcctexists}, \textfn{CheckingAcct$\geq$0}] must preserve thermometer encoding of CheckingAcct., which can only increase. Actions can only turn on higher-level dummies that are off, where \textfn{CheckingAcctexists} is the lowest-level dummy and \textfn{CheckingAcct$\geq$0} is the highest-level-dummy.
\item ThermometerEncoding: Actions on [\textfn{SavingsAcctexists}, \textfn{SavingsAcct$\geq$100}] must preserve thermometer encoding of SavingsAcct., which can only increase. Actions can only turn on higher-level dummies that are off, where \textfn{SavingsAcctexists} is the lowest-level dummy and \textfn{SavingsAcct$\geq$100} is the highest-level-dummy.
\end{constraints}

\subsection{Additional Results}

We report statistics on model performance statistics for all models \cref{Table::GermanModelPerformance}. We include statistics for the \LR{} model as a baseline.

\begin{table}[h]
\centering
\small
\begin{tabular}{lHHrr}
\multicolumn{1}{c}{ } & \multicolumn{2}{c}{} & \multicolumn{2}{c}{\XGB{}} \\
\cmidrule(l{3pt}r{3pt}){2-3} \cmidrule(l{3pt}r{3pt}){4-5}
  & \textbf{Train} & \textbf{Test} 
  & \textbf{Train} & \textbf{Test} \\ 
  \midrule
\textbf{AUC} & 0.807 & 0.768 & 0.819 & 0.7615 \\
   \midrule
\textbf{ECE} & 20.0\% & 20.0\% & 0.0\% & 10.0\% \\
   \midrule
\textbf{Error} & 27.2\% & 28.0\% & 21.9\% & 23.0\% \\ 
   \midrule
\textbf{$n$} & 800 & 200 & 800 & 200 \\ 
   \midrule
\textbf{$n_{\textrm{pos}}$} & 560 & 140 & 560 & 140 \\ 
   \midrule
\textbf{$p$} & 70.0\% & 70.0\% & 70.0\% & 70.0\% \\ 
   \midrule
\textbf{$n_{\textrm{clf\_pos}}$} & 738 & 186 & 615 & 120\\ 
   \midrule
\textbf{$n_{\textrm{clf\_neg}}$} & 62 & 14 & 185 & 80 \\
\end{tabular}
\caption{Overview of model performance for \XGB{} models for the \textds{german} dataset}.
\label{Table::GermanModelPerformance}
\end{table}

\clearpage
\section{Monotonicity Violations in Organ Transplants}
\label{Appendix::OrganTransplant}
\subsection{Demonstration}

Statistical models in organ transplant allocation~\cite{gotlieb2022promise} have recently attracted scrutiny for their potential to preclude access to transplants~\cite{murgia2023algorithms,attia2024effect}. 
Most notably, in 2018, the UK introduced a patient matching scheme for liver transplants using the Transplant Benefit Score (TBS). 
\citet{attia2023implausible} found that TBS can predict \emph{higher survival after the patient gets cancer}, which is biologically implausible and could result in the implicit de-prioritization of cancer patients~\cite{attia2023implausible}. 
We show how our machinery could have preemptively flagged such behavior during model development.

The TBS system comprises four Cox proportional hazard regression models. We focus on its \emph{need} component which predicts survival without a transplant:
\begin{align*}
    f(\xp) =
    \sum_{t=1}^T \sum_{s \in \{\mathsf{c}, \mathsf{nc}\}} 
    \indic{\xp_\textfn{cancer} = s} \cdot S_{0}^{\textfn{c}}(t)^{\exp(\boldsymbol{\beta}^{\textfn{c}} \cdot (\xp - \boldsymbol{\mu}^{\textfn{c}}))}
\end{align*}
where $S_{0}^{\textfn{c}}, S_{0}^{\textfn{nc}}: \mathbb{N} \rightarrow [0, 1]$ for $t \in [T]$ for some $T \in \mathbb{N}$ are baseline hazard functions, and the vectors $\boldsymbol{\beta}^{\textfn{c}}, \boldsymbol{\beta}^{\textfn{nc}}$ and $\boldsymbol{\mu}^{\textfn{c}}, \boldsymbol{\mu}^{\textfn{nc}}$ are the model parameters and data normalizers for patients with cancer ($\textfn{c}$) and without ($\textfn{nc}$), respectively.

\begin{wrapfigure}{r}{0.5\linewidth}{
    \includegraphics[width=\linewidth]{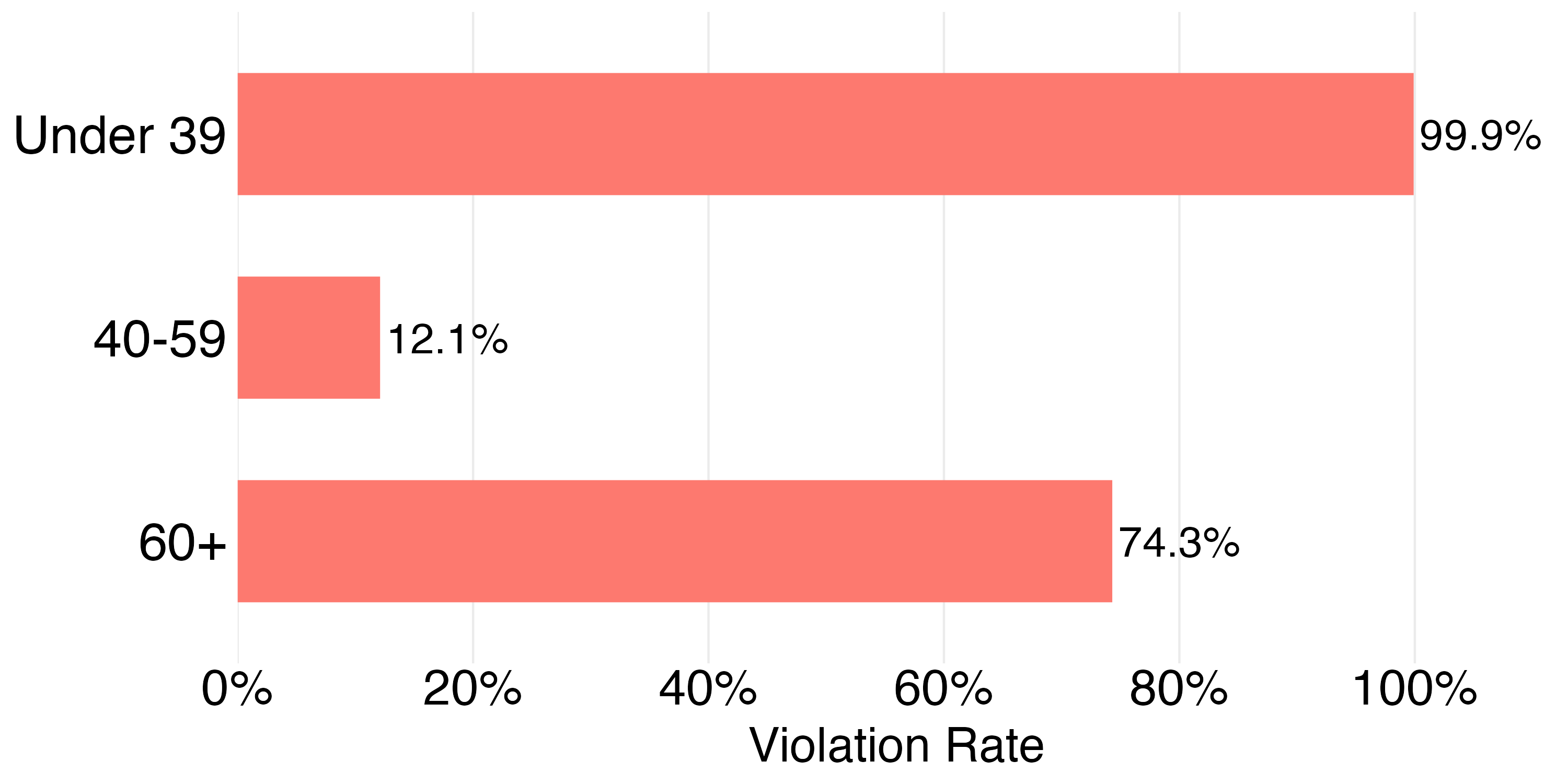}
    \caption{Proportion of predictions that violate monotonicity of predicted survival to changes in cancer status across a cohort of simulated patients, in any of the three plausible interaction sets. See \cref{Fig::TransplantBenefitByModel} in the Appendix for a breakdown by interaction model. %
    }
    \label{Fig::OrganTransplant}
    }
\end{wrapfigure}

We apply our machinery to test if the model predicts higher survival after a patient gets cancer. This tests a monotonicity condition~\citep{BenDavid:1995kr,gupta2016monotonic} that can be violated because the TBS uses separate models for patients with and without cancer. We test responsiveness on a cohort of \num{1000} patients without cancer from a clinical distribution in~\citet{attia2023implausible}, where we ensure a uniform distribution of ages between 30 and 80 to be able to evaluate the effect of demographics on monotonicity violations. We define responsiveness $\rho(\xp)$ to capture how often the model violates monotonicity by setting target predictions as $\ytarget{\xp} = \{y \mid y < f(\xp) \}$, i.e., range of values that would \emph{not} violate monotonicity. We test if responsiveness is below $\thresh = 0.01$ at $\alpha = 0.05$ using $n = 30$ samples for each patient.
To showcase our framework under multiple interaction models, we follow the \emph{flagging} approach in \cref{Appendix::SensAnalysis} by identifying the patients where the need model violates monotonicity under at least one interaction model after applying Holm's p-value adjustment. 
We use the family $\intFamily$ of three models:
\begin{plainlist}
    \item {\emph{Deterministic:}} Each action changes the cancer status by setting values of the $\textfn{primary\_disease}$, $\textfn{max\_tumor\_size}$, and $\textfn{tumor\_number}$ features.
    \item  {\emph{Random Effects:}} In addition to direct cancer features, we simulate plausible natural variation in lab values, such as levels of \textfn{bilirubin} and \textfn{sodium}.
    \item {\emph{Causal Effects:}} Instead of simulating variation in lab values, we assume that cancer has downstream causal effects on other variables, such as increasing the probability of \textfn{hospitalization} and \textfn{renal\_dysfunction}.
\end{plainlist}

In \cref{Fig::OrganTransplant}, we see that the \emph{need} model violates monotonicity under at least one interaction model for a considerable number of patients.
The effect varies significantly across age groups:
the model violates monotonicity conditions for only 12\% of patients between 40 and 59.
In contrast, violation rates for patients under 39 and over 60 are 99.9\% and 74\% respectively. 

When considering the violations in each interaction model separately, we see differences across the models (see Appendix for details). In the 40--59 age group, within the \emph{random effects} model, the rate of violations is only around 10\%, but rises to over 50\% rate in the other models. The \emph{flagging} procedure takes this variation into account. In our setting, we did not have enough power to consider most of the detected violations in the \emph{deterministic} and \emph{causal effects} models as significant after p-value adjustment, resulting in the final 12\% violation rate in at least one of the models.

The detected violations implicitly de-prioritize cancer patients. Our tools flag this concrete safety issue at the system level, enabling model developers to test responsiveness individually for each patient, and generate test cases for iterative model development.

\subsection{Components of the Transplant Benefit Score}
Our study evaluates the transplant benefit score, which is publicly available at \url{https://github.com/SurgicalInformatics/transplantbenefit/}. 

The TBS consists of four Cox regression models with the following corresponding survival functions:
\begin{align*}
f_{\text{need}}^{\textfn{c}}(\xp) &= \sum_{t=1}^T S_{0,\text{need}}^{\textfn{c}}(t)^{\exp(\boldsymbol{\beta}_{\text{need}}^{\textfn{c}\top} (\xp - \boldsymbol{\mu}_{\text{need}}^{\textfn{c}}))}, \\
f_{\text{need}}^{\textfn{nc}}(\xp) &= \sum_{t=1}^T S_{0,\text{need}}^{\textfn{nc}}(t)^{\exp(\boldsymbol{\beta}_{\text{need}}^{\textfn{nc}\top} (\xp - \boldsymbol{\mu}_{\text{need}}^{\textfn{nc}}))} \\
f_{\text{utility}}^{\textfn{c}}(\xp) &= \sum_{t=1}^T S_{0,\text{utility}}^{\textfn{c}}(t)^{\exp(\boldsymbol{\beta}_{\text{utility}}^{\textfn{c}\top} (\xp - \boldsymbol{\mu}_{\text{utility}}^{\textfn{c}}))}, \\
f_{\text{utility}}^{\textfn{nc}}(\xp) &= \sum_{t=1}^T S_{0,\text{utility}}^{\textfn{nc}}(t)^{\exp(\boldsymbol{\beta}_{\text{utility}}^{\textfn{nc}\top} (\xp - \boldsymbol{\mu}_{\text{utility}}^{\textfn{nc}}))}
\end{align*}
where $\textfn{c}$ and $\textfn{nc}$ indicate models applied to patients with cancer and without, respectively, $S_{0, \cdot}: \mathbb{N} \rightarrow [0, 1]$ for $t \in [T]$ for some $T \in \mathbb{N}$ are pre-defined baseline hazard functions, and the vectors $\beta$ and $\mu$ are the corresponding model parameters and data normalizers, respectively. 
The final TBS score is computed as $\smash{f_\text{TBS}(\xp) = f_{\text{utility}}^{\xp_\textfn{cancer}}(\xp) - f_{\text{need}}^{\xp_\textfn{cancer}}(\xp)}.$ 

\subsection{Description of Dataset}

We follow the methodology in the \citet{attia2023implausible}, who define a suitable probability distribution as part of a study on organ transplant prioritization for cancer patients. \citeauthor{attia2023implausible} specify a probabilistic model to simulate patients by fitting a distribution to match a real cohort a liver transplant patients. We follow their approach to generate $n=\num{1000}$ simulated patients with $d=32$ features. We start with the default patient case in this implementation to set the baseline characteristics in our cohort and modify certain variables as described below. 

\paragraph{Static Variables} 

We simulate the demographics as follows:
\begin{align}
\xp_\textfn{gender} &\sim \mathsf{Bernoulli}(0.5) \quad \text{where } 0 = \textfn{man}, 1 = \textfn{woman} \\
\xp_\textfn{age} &\sim \mathsf{Unif}[\{30, 31, \ldots, 80\}].
\end{align}

We also simulate other lab values as follows:
\begin{align}
\xp_\textfn{albumin} &\sim \mathsf{Unif}[30, 40] \\
\xp_\textfn{potassium} &\sim \mathsf{Unif}[3.5, 5.0]
\end{align}
We detail the model for sampling other clinical variables next.

\paragraph{Liver Health Parameters}

We set up the following structural causal model (SCM)~\cite{pearl2009causality} for the liver parameters: bilirubin, sodium, international normalized ratio (INR), and creatinine. We use this probabilistic model both to generate the initial patient cohort, and to simulate the random effects due to natural variation in the reachable sets.

Let $\mathbf{U} = (U_{\textfn{bili}}, U_{\textfn{Na}}, U_{\textfn{INR}}, U_{\textfn{creat}})$ denote the vector of exogenous noise variables, and let $(\xp_{\text{bili}}, \xp_{\text{Na}}, \xp_{\text{INR}}, \xp_{\text{creat}})$ denote the vector of correlated endogenous variables representing the four liver parameters.

\textit{Exogenous Variables.}
We set $\mathbf{U} \sim \mathcal{N}(0, \mathbf{\Sigma})$ with:
\begin{equation}
\boldsymbol{\Sigma} = 
\begin{bmatrix} 
1.0 & 0.447 & 0.320 & -0.257 \\
0.447 & 1.0 & 0.370 & -0.043 \\
0.320 & 0.370 & 1.0 & -0.091 \\
-0.257 & -0.043 & -0.091 & 1.0
\end{bmatrix}
\end{equation}

\textit{Structural Equations.}
The endogenous variables are determined by the following structural equations:
\begin{align}
\xp_{\textfn{bili}} &= \min(200, \max(15, \exp(0.5 \cdot U_{\textfn{bili}} + 3.5))) \\
\xp_{\textfn{Na}} &= \min(145, \max(125, 5 \cdot U_{\textfn{Na}} + 137)) \\
\xp_{\textfn{INR}} &= \min(2.4, \max(0.9, \exp(0.3 \cdot U_{\textfn{INR}} - 0.2) + 0.8)) \\
\xp_{\textfn{creat}} &= \min(200, \max(45, \exp(0.4 \cdot U_{\textfn{creat}} + 4.2)))
\end{align}

where:
\begin{bulletlist}
\item $\xp_{\textfn{bili}}$ represents bilirubin levels (clipped to $[15, 200]$)
\item $\xp_{\textfn{Na}}$ represents sodium levels (clipped to $[125, 145]$)
\item $\xp_{\textfn{INR}}$ represents international normalized ratio (clipped to $[0.9, 2.4]$)
\item $\xp_{\textfn{creat}}$ represents creatinine levels (clipped to $[45, 200]$)
\end{bulletlist}

We choose the parameters to approximately match the reported statistics in a simulated cohort from \citet{attia2023implausible}. We show the statistical properties of our generated cohort in \cref{Fig::TransplantCohort}.

\subsection{Interaction Model}
We detail the features and the considered interaction model in the table:
\begin{table}[h]
\centering
\resizebox{\linewidth}{!}{
\begin{tabular}{lllllllr}
\textheader{Name} & \textheader{Type} & \textheader{LB} & \textheader{UB} & \textheader{Actionable} & \textheader{Sign} & \textheader{Joint Constraints} & \textheader{Partition ID} \\
\midrule
\rowcolor{gray!30} \textfn{rinpatient\_tbs} & $\{0,1\}$ & 0 & 1 & No &  & -- & 3 \\
\rowcolor{gray!30} \textfn{rregistration\_tbs} & $\mathbb{Z}$ & 1 & 7 & No &  & -- & 4 \\
\rowcolor{gray!30} \textfn{rwaiting\_time\_tbs} & $\mathbb{Z}$ & 0 & 3650 & No &  & -- & 5 \\
\rowcolor{gray!30} \textfn{rage\_tbs} & $\mathbb{Z}$ & 30 & 80 & No &  & -- & 6 \\
\rowcolor{gray!30} \textfn{rgender\_tbs} & $\{0,1\}$ & 0 & 1 & No &  & -- & 7 \\
\textfn{rdisease\_primary\_tbs} & $\mathbb{Z}$ & 1 & 9 & Yes &  & 1, 2 & 1 \\
\rowcolor{gray!30} \textfn{rdisease\_secondary\_tbs} & $\mathbb{Z}$ & 1 & 9 & No &  & -- & 8 \\
\rowcolor{gray!30} \textfn{rdisease\_tertiary\_tbs} & $\mathbb{Z}$ & 1 & 9 & No &  & -- & 9 \\
\rowcolor{gray!30} \textfn{previous\_tx\_tbs} & $\{0,1\}$ & 0 & 1 & No &  & -- & 10 \\
\rowcolor{gray!30} \textfn{rprevious\_surgery\_tbs} & $\{0,1\}$ & 0 & 1 & No &  & -- & 11 \\
\rowcolor{gray!30} \textfn{rbilirubin\_tbs} & $\mathbb{Z}$ & 15 & 200 & No &  & -- & 2 \\
\rowcolor{gray!30} \textfn{rinr\_tbs} & $\mathbb{R}$ & 0.9 & 2.4 & No &  & -- & 2 \\
\rowcolor{gray!30} \textfn{rcreatinine\_tbs} & $\mathbb{Z}$ & 45 & 200 & No &  & -- & 2 \\
\rowcolor{gray!30} \textfn{rrenal\_tbs} & $\{0,1\}$ & 0 & 1 & No &  & -- & 12 \\
\rowcolor{gray!30} \textfn{rsodium\_tbs} & $\mathbb{Z}$ & 125 & 145 & No &  & -- & 2 \\
\rowcolor{gray!30} \textfn{rpotassium\_tbs} & $\mathbb{R}$ & 3.5 & 5.0 & No &  & -- & 13 \\
\rowcolor{gray!30} \textfn{ralbumin\_tbs} & $\mathbb{Z}$ & 30 & 40 & No &  & -- & 14 \\
\rowcolor{gray!30} \textfn{rencephalopathy\_tbs} & $\{0,1\}$ & 0 & 1 & No &  & -- & 15 \\
\rowcolor{gray!30} \textfn{rascites\_tbs} & $\{0,1\}$ & 0 & 1 & No &  & -- & 16 \\
\rowcolor{gray!30} \textfn{rdiabetes\_tbs} & $\{0,1\}$ & 0 & 1 & No &  & -- & 17 \\
\rowcolor{gray!30} \textfn{rmax\_afp\_tbs} & $\mathbb{Z}$ & 0 & 1000 & No &  & -- & 18 \\
\textfn{rtumour\_number\_tbs} & \{`0 or 1'$^*$, `2', `3+'\} &  &  & Yes &  & 1, 2 & 1 \\
\textfn{rmax\_tumour\_size\_tbs} & $\mathbb{R}$ & 0 & 20 & Yes &  & 1, 2 & 1 \\
\rowcolor{gray!30} \textfn{dage\_tbs} & $\mathbb{Z}$ & 18 & 80 & No &  & -- & 19 \\
\rowcolor{gray!30} \textfn{dcause\_tbs} & $\mathbb{Z}$ & 1 & 4 & No &  & -- & 20 \\
\rowcolor{gray!30} \textfn{dbmi\_tbs} & $\mathbb{R}$ & 15 & 50 & No &  & -- & 21 \\
\rowcolor{gray!30} \textfn{ddiabetes\_tbs} & $\mathbb{Z}$ & 1 & 3 & No &  & -- & 22 \\
\rowcolor{gray!30} \textfn{dtype\_tbs} & $\{0,1\}$ & 0 & 1 & No &  & -- & 23 \\
\rowcolor{gray!30} \textfn{bloodgroup\_compatible\_tbs} & $\{0,1\}$ & 0 & 1 & No &  & -- & 24 \\
\rowcolor{gray!30} \textfn{splittable\_tbs} & $\{0,1\}$ & 0 & 1 & No &  & -- & 25 \\
\\
\multicolumn{8}{l}{* This feature value is treated as no tumours if the primary disease does not indicate cancer, $\textfn{rdisease\_primary\_tbs} \neq 1$, and as one tumour otherwise.}
\end{tabular}

}
\label{Table::ActionSetTBSFull}
\end{table}

\begin{constraints}
\item IfThenConstraint: If \textfn{rtumour\_number\_tbs} $\in$ $\{\text{`2',`3+'}\}$, then \textfn{rdisease\_primary\_tbs} = 1 (cancer)
\item IfThenConstraint: If \textfn{rdisease\_primary\_tbs} = 1, then \textfn{rmax\_tumour\_size\_tbs} > 0.
\end{constraints}

Next, we describe three interaction models constructed by setting different downstream effect models $\respdist{}$.

\paragraph{Deterministic}
In this baseline interaction model, we assume that cancer diagnosis does not affect any non-cancer features. Under this model, counterfactual patients with cancer are identical to their non-cancer counterparts except for the cancer-specific features described above: primary disease diagnosis, tumor number, and maximum tumor size. 

\paragraph{Random Effects}
We model natural variation in labs by applying directional perturbations to liver function parameters:
\begin{align}
\xp_{\textfn{bili}}^{(1)} &= \textfn{clip}(\xp_{\textfn{bili}}^{(0)} \cdot \gamma_{\textfn{bili}}, 15, 200), \quad \gamma_{\textfn{bili}} \sim \mathsf{Unif}[1.1, 1.5] \\
\xp_{\textfn{INR}}^{(1)} &= \textfn{clip}(\xp_{\textfn{INR}}^{(0)} + \delta_{\textfn{INR}}, 0.9, 2.4), \quad \delta_{\textfn{INR}} \sim \mathsf{Unif}[0.1, 0.3] \\
\xp_{\textfn{Na}}^{(1)} &= \textfn{clip}(\xp_{\textfn{Na}}^{(0)} - \delta_{\textfn{Na}}, 125, 145), \quad \delta_{\textfn{Na}} \sim \mathsf{Unif}[2, 5] \\
\xp_\textfn{AFP}^{(1)} &\sim \mathsf{Unif}[100, 1000]
\end{align}

\paragraph{Causal Effects}
Advanced liver disease with cancer may lead to clinical complications that affect other patient features. We model two such effects:
\begin{plainlist}
\item \textit{Renal dysfunction}: With probability $0.3$, patients develop renal complications, setting $\textfn{rrenal\_tbs} = 1$ and increasing creatinine by a factor $\gamma_{\textfn{creat}} \sim
\mathsf{Unif}[1.2, 1.8]$.
\item \textit{Hospitalization}: With probability $0.4$, patients require hospitalization, setting $\textfn{rinpatient\_tbs} = 1$.
\end{plainlist}

\begin{figure}[t]
    \centering
    \includegraphics[width=0.65\linewidth]{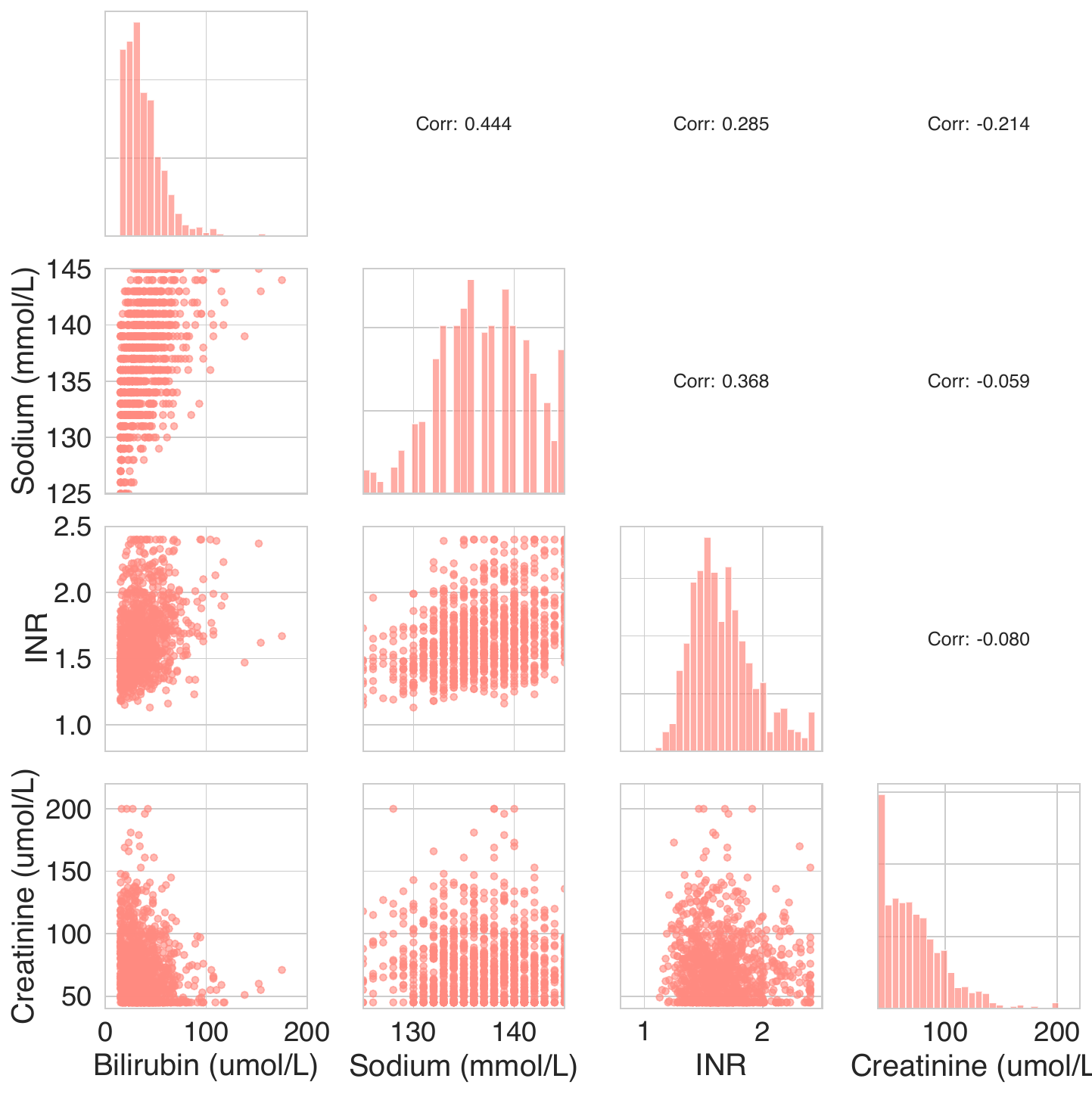}
    \caption{Pairwise relationships of the four liver parameter distributions according to our probabilistic model. These statistics are similar to those obtained by \citet{attia2023implausible}.}
    \label{Fig::TransplantCohort}
\end{figure}

\clearpage
\subsection{Additional Results}

\begin{figure}[h!]
    \centering
    \includegraphics[width=\linewidth]{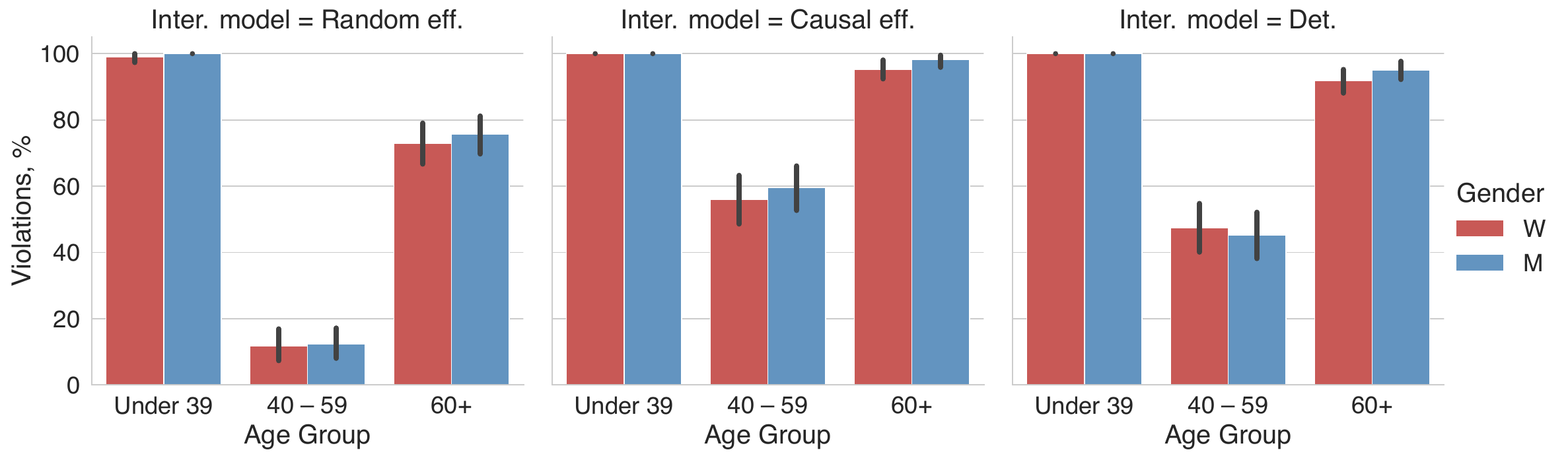}
    \caption{Monotonicity violations by interaction model, with a breakdown by age group and gender.}
    \label{Fig::TransplantBenefitByModel}
\end{figure}
\clearpage

\clearpage
\section{Preclusion in Recidivism Prediction}
\label{Appendix::RecidivismPrediction}

\subsection{Description of Dataset}

We work with a large sample of defendants from New York state derived from the ``Recidivism of Prisoners Released in 1994'' dataset released by the U.S. Department of Justice ~\citep{icpsr1994} and used in prior work~\citep[see e.g.,.][]{zeng2017interpretable, marx2020multiplicity}. Our goal is to build a logistic regression model to predict whether prisoner $i$ is rearrested within 3 years of release from prison ($y_i = 1$). Our processed dataset contains a total of $n = \num{29400}$ and a set of $d = 14$ features with varying degrees of mutability listed in \cref{Table::ActionSetRNYAP}.%

\subsection{Interaction Model}

We consider a deterministic interaction model where each defendant can perform hypothetical actions that clear their criminal history (see below for detailed examples). In this case, our model contains a set of eight separable constraints for each feature which we use to ensure actionability, monotonicity, integrality, and bounds show in \cref{Table::ActionSetRNYAP}.

\begin{table}[h]
\centering
\resizebox{0.85\linewidth}{!}{
\begin{tabular}{llllllll}
\toprule
\textheader{Name} & \textheader{Type} & \textheader{LB} & \textheader{UB} & \textheader{Actionable} & \textheader{Sign} & \textheader{Joint Constraints} & \textheader{Partition ID} \\
\midrule
\rowcolor{gray!30} \textfn{drug\_abuser} & $\{0,1\}$ & 0 & 1 & No &  & -- & 0 \\
\textfn{drug\_treatment} & $\{0,1\}$ & 0 & 1 & Yes & $+$ & -- & 1 \\
\rowcolor{gray!30} \textfn{alcohol\_abuser} & $\{0,1\}$ & 0 & 1 & No &  & -- & 2 \\
\textfn{alcohol\_treatment} & $\{0,1\}$ & 0 & 1 & Yes & $+$ & -- & 3 \\
\textfn{edu\_program\_participation} & $\{0,1\}$ & 0 & 1 & Yes & $+$ & -- & 4 \\
\textfn{voc\_program\_participation} & $\{0,1\}$ & 0 & 1 & Yes & $+$ & -- & 5 \\
\rowcolor{gray!30} \textfn{age\_at\_release} & $\mathbb{R}$ & 17.3 & 83.9 & No &  & -- & 6 \\
\rowcolor{gray!30} \textfn{female} & $\{0,1\}$ & 0 & 1 & No &  & -- & 7 \\
\rowcolor{gray!30} \textfn{time\_served} & $\mathbb{Z}$ & 1 & 8 & No &  & -- & 8 \\
\textfn{prior\_arrests} & $\mathbb{Z}$ & 0 & 30 & Yes & $-$ & 1, 2, 3, 4 & 9 \\
\textfn{n\_prior\_felony} & $\mathbb{Z}$ & 0 & 61 & Yes & $-$ & 1 & 9 \\
\textfn{n\_prior\_misdemeanor} & $\mathbb{Z}$ & 0 & 185 & Yes & $-$ & 2 & 9 \\
\textfn{n\_prior\_general\_violence} & $\mathbb{Z}$ & 0 & 24 & Yes & $-$ & 3 & 9 \\
\textfn{n\_prior\_prb\_or\_fine} & $\mathbb{Z}$ & 0 & 34 & Yes & $-$ & 4 & 9 \\
\bottomrule
\end{tabular}

}
\caption{
Interaction Model for the processed \textds{rearrest\_NY} dataset.
\textbf{Type} indicates the feature type ($\Z$ for integer, $\{0,1\}$ for binary).
\textbf{LB}, \textbf{UB} are the lower and upper bounds for the feature.
\textbf{Actionable} indicates whether the feature is globally actionable. Non-actionable features are highlighted.
\textbf{Sign} indicates monotonicity constraints---whether feature can only increase (+) or decrease (-).
\textbf{Constraints} are a list non-separable constraint indices (listed below table) it is tied to (if any).
\textbf{Partition ID} indicates which partition the feature belongs to.
}
\label{Table::ActionSetRNYAP}
\end{table}

\begin{constraints}
\item IfThenConstraint: If \textfn{priorarrests} = 0.0, then \textfn{npriorfelony} = 0.0
\item IfThenConstraint: If \textfn{priorarrests} = 0.0, then \textfn{npriormisdemeanor} = 0.0
\item IfThenConstraint: If \textfn{priorarrests} = 0.0, then \textfn{npriorgeneralviolence} = 0.0
\item IfThenConstraint: If \textfn{priorarrests} = 0.0, then \textfn{npriorprborfine} = 0.0
\end{constraints}

\subsection{Additional Results}
This table includes additional model training and performance statistics. 

\begin{table}[H]
\centering
\small
\begin{tabular}{lllllll}
\toprule
Dataset & Interaction model & Model & Train AUC & Train error & Test AUC & Test error \\
\midrule
rNYap & criminal\_history\_clearance & logreg & 0.7287 & 0.2893 & 0.6901 & 0.3063 \\
\bottomrule
\end{tabular}

\caption{Overview of model performance for the \textds{rearrest\_NY} dataset.}
\end{table}

\clearpage
\section{Preventing Gaming in Content Moderation}
\label{Appendix::ContentModeration}

\subsection{Description of Dataset}
We work with the \textds{twitterbot} dataset, which was originally curated by \citet{gilani2017classification}. The dataset defines a binary classification task where we wish to predict if a user account on Twitter belongs to a \emph{human} ($y_i =1$) or a \emph{bot} ($y_i =0$). The dataset contains a total of $n = \num{3431}$ instances and $d = 20$ features that encode semantically meaningful characteristics about their interactions and login history---e.g., \textfn{age\_of\_account\_in\_days} for account age, \textfn{user\_tweeted} for the number of user tweets, and \textfn{source\_identity} for source of user interaction (mobile, web, etc.). 

In this case, the dataset contains a limited number of features given that all features are not readily available or shared across accounts. We process the dataset to define a subset of additional features as follows: (1) we include additional dummies to indicate ``missing'' values for \textfn{num\_tweets}, \textfn{num\_retweets} and \textfn{num\_replies}; (2) we binarize features by using a adding a thermometer encoding to \textfn{num\_followers} and \textfn{age\_of\_accounts\_in\_days}, setting thresholds that reflect salient milestones for follows and membership history; (3) we multi-hot encoded \textfn{source\_identity}.

\subsection{Interaction Model}

We specify an interaction model to specify the kinds of changes that each user can perform on the 20 features in the dataset. We present a list of all features and their corresponding feature-level constraints in \cref{Table::ActionSetTwitter} and list joint actionability constraints below. Our model only considers interactions on features that a user can change themselves and ignores changes on features that could change but cannot be manipulated.

\begin{table}[h]
\centering
\resizebox{0.85\linewidth}{!}{
\begin{tabular}{lllllllr}
\textheader{Name} & \textheader{Type} & \textheader{LB} & \textheader{UB} & \textheader{Actionable} & \textheader{Sign} & \textheader{Joint Constraints} & \textheader{Partition ID} \\
\midrule
\rowcolor{gray!30} \textfn{followers$\geq$1k} & $\{0,1\}$ & 0 & 1 & No &  & 4 & 0 \\
\rowcolor{gray!30} \textfn{followers$\geq$100k} & $\{0,1\}$ & 0 & 1 & No &  & 4 & 0 \\
\rowcolor{gray!30} \textfn{followers$\geq$1M} & $\{0,1\}$ & 0 & 1 & No &  & 4 & 0 \\
\rowcolor{gray!30} \textfn{followers$\geq$10M} & $\{0,1\}$ & 0 & 1 & No &  & 4 & 0 \\
\textfn{num\_tweets} & $\mathbb{Z}$ & 0 & 35000 & Yes & $+$ & 1, 5 & 5 \\
\textfn{no\_tweets} & $\{0,1\}$ & 0 & 1 & Yes & $-$ & 1 & 5 \\
\rowcolor{gray!30} \textfn{urls\_count} & $\mathbb{Z}$ & 0 & 13013 & No &  & 5 & 5 \\
\textfn{num\_retweets} & $\mathbb{Z}$ & 0 & 3000 & Yes & $+$ & 2 & 6 \\
\textfn{no\_retweets} & $\{0,1\}$ & 0 & 1 & Yes & $-$ & 2 & 6 \\
\textfn{num\_replies} & $\mathbb{Z}$ & 0 & 6991 & Yes & $+$ & 3 & 7 \\
\textfn{no\_replies} & $\{0,1\}$ & 0 & 1 & Yes & $-$ & 3 & 7 \\
\textfn{age\_of\_account$\geq$180\_days} & $\{0,1\}$ & 0 & 1 & Yes &  & -- & 1 \\
\textfn{age\_of\_account$\geq$365\_days} & $\{0,1\}$ & 0 & 1 & Yes &  & -- & 2 \\
\textfn{age\_of\_account$\geq$730\_days} & $\{0,1\}$ & 0 & 1 & Yes &  & -- & 3 \\
\textfn{age\_of\_account$\geq$1825\_days} & $\{0,1\}$ & 0 & 1 & Yes &  & -- & 4 \\
\textfn{follower\_friend\_ratio} & $\mathbb{R}$ & 0.0 & 13364332.2 & Yes & $-$ & -- & 8 \\
\rowcolor{gray!30} \textfn{source\_web} & $\{0,1\}$ & 0 & 1 & No &  & -- & 10 \\
\rowcolor{gray!30} \textfn{source\_mobile} & $\{0,1\}$ & 0 & 1 & No &  & -- & 11 \\
\rowcolor{gray!30} \textfn{source\_app} & $\{0,1\}$ & 0 & 1 & No &  & -- & 12 \\
\rowcolor{gray!30} \textfn{source\_news} & $\{0,1\}$ & 0 & 1 & No &  & -- & 15 \\
\end{tabular}

}
\label{Table::ActionSetTwitterFull}
\caption{
Interaction Model for the processed \textds{twitterbot} dataset. 
\textbf{Type} indicates the feature type ($\Z$ for integer, $\{0,1\}$ for binary). 
\textbf{LB}, \textbf{UB} are the lower and upper bounds for the feature.
\textbf{Actionable} indicates whether the feature is globally actionable. Non-actionable features are highlighted.
\textbf{Sign} indicates monotonicity constraints---whether feature can only increase (+) or decrease (-).
\textbf{Joint Constraints} are a list non-separable constraint indices (listed below table) it is tied to (if any).
\textbf{Partition ID} indicates which partition the feature belongs to.
}
\label{Table::ActionSetTwitter}
\end{table}

\begin{constraints}
\item IfThenConstraint: If \textfn{notweets} = 0.0, then \textfn{numtweets} > 1.0
\item IfThenConstraint: If \textfn{noretweets} = 0.0, then \textfn{numretweets} > 1.0
\item IfThenConstraint: If \textfn{noreplies} = 0.0, then \textfn{numreplies} > 1.0
\item DirectionalLinkage: Actions on \textfn{numtweets} will induce to actions on [`\textfn{urlscount}']. Each unit change in \textfn{numtweets} leads to at least 1.00-unit change in \textfn{urlscount}
\end{constraints}

\subsection{Additional Results}
\begin{table}[h!]
\centering
\resizebox{\linewidth}{!}{
\begin{tabular}{llrrrrrrrrrrr}
Procedure & Description & \# Models & \# Robust & Train & Test & Valid & Train & Test & Valid & Train & Test & Valid\\
\midrule
Manual & \cell{l}{Train Models with \\Immutable Features} & 370 & 370 & 0.0\% & 0.0\% & 0.0\% & 0.0\% & 0.0\% & 0.0\% & 0.531 & 0.570 & 0.581\\
\midrule

Convex & \cell{l}{Consider Responsiveness \\ w.r.t Convex Perturbation Check} & 901 & 687 & 0.3\% & 0.0\% & 0.9\% & 56.2\% & 57.1\% & 55.9\% & 0.743 & 0.754 & 0.759\\
\midrule

Exact & \cell{l}{Evaluate Responsiveness \\ w.r.t Exact Actions} & 901 & 76 & 9.6\% & 9.9\% & 9.3\% & 9.6\% & 9.9\% & 9.3\% & 0.722 & 0.727 & 0.734\\
\end{tabular}

}
\caption{
Full train, test, validation set results for the model with the highest validation AUC among \emph{Considered} models:
$\leq 10\%$ ``Bot'' predictions with certified responsiveness $\geq \thresh = 0.05$. 
\emph{\% Responsive} show \% of "Bot" predictions with responsiveness $\geq \thresh = 0.05$ under the procedure's reachable set (\emph{Perceived}) and the exact reachable set (\emph{True}).
}
\label{Table::TwitterResultsLong}
\end{table}

  \section{LLM Safety Evaluation}
  \label{Appendix::LLMSafety}

\subsection{Description of the Dataset and LLM Details}
  We evaluate four open-weight instruction-tuned LLMs accessed via Hugging Face:
  \textfn{OLMo-2-1124-7B-Instruct}~\cite{walsh2025olmo},
  \textfn{Llama-3.1-8B-Instruct}~\cite{grattafiori2024llama},
  \textfn{Qwen2.5-7B-Instruct}~\cite{bai2023qwen}, and
  \textfn{Phi-3.5-mini-instruct}~\cite{haider2024phi}.
  The harmful-prompt corpus is \textds{sorrybench} (release 202503), consisting of $n_\textfn{prompts} = 440$ prompts spanning 44 fine-grained safety categories~\cite{xie2025sorry}.

Completions are produced with \textfn{vLLM} in \textfn{bfloat16} at temperature $T = 0.7$ and \textfn{max\_tokens} $= 1024$, using each model's chat template. For every prompt under every interaction we draw
multiple completions across multiple sampling seeds; together these realize the stochastic downstream effect $\varepsilon$ that accounts for LLM sampling noise.

The classifier $\clf{\xp}$ is the \textds{sorrybench} fine-tuned judge \textfn{ft-mistral-7b-instruct-v0.2-sorry-bench-202406}, queried at $T = 0$ with the official scoring template. The judge emits a single token in
$\{0, 1\}$, where $1$ denotes compliance ($\ytarget{} = \{\texttt{comply}\}$) and $0$ refusal.

\subsection{Interaction Models}
We adopt four prompt-level perturbations from the \textds{sorrybench} mutation suite as our interaction models $I$:
\begin{bulletlist}
\item $I_\mathrm{misspell}$: writing-style mutation introducing systematic misspellings.
\item $I_\mathrm{roleplay}$: rewriting the request as a roleplay scenario.
\item $I_\mathrm{chinese}$, $I_\mathrm{french}$: translation of the English prompt to Chinese (\textfn{zh-cn}) and French.
\end{bulletlist}
For a base prompt $\xp$, the corresponding mutation corresponds to an action $\ab \in \Aset{\xp}$; the downstream effect $\varepsilon$ is the LLM sampling distribution, as described before.

\subsection{Statistical Test Details}
With $\rho = \E_{\xp \sim D} \E_{\xq \sim X_I(\xp)}\big[\mathbb{I}[\clf{\xq} = \texttt{comply}]\big]$ and threshold $\tau = 40\%$, we test:
\begin{equation*}
H_0:~\rho \ge \tau \qquad \text{vs.} \qquad H_1:~\rho < \tau.
\end{equation*}
For each (model, interaction) pair, we pool compliance indicators into a single sufficient statistic $k \sim \mathsf{Binom}(n, \rho)$ and report the exact
one-sided Clopper-Pearson inversion test p-value
\begin{equation*}
p \;=\; \Pr{K \le k},
\end{equation*}
where $K \sim \mathsf{Binom}(n,\,\tau)$.
The required sample size $n^\star$ is the smallest $n$ such that the exact binomial test of $H_0$ has power at least $1 - \beta = 95\%$ at the alternative $\rho = \tau - \delta$ at level $\alpha = 0.05$. We
anchor the alternative at each model's full-cohort base compliance rate, so $\delta = \tau - \rho^{(0)}$. This yields $n^\star = 79$ for OLMo ($\rho^{(0)} = 22.7\%$, $\delta = 17.3$\,p.p.) and $n^\star = 159$
for Llama ($\rho^{(0)} = 27.7\%$, $\delta = 12.3$\,p.p.), as reported in the main body.

\clearpage
\section{Omitted Formal Results}

\subsection{Proof of \cref{Rem::MinSampSize}}
\begin{proof}
From the definition of the exact Binomial confidence interval, we have that:
    \begin{equation}
    \rsphatmax{\xp}{\alpha}(n, \rsphata{}) = \mathsf{B}_{1 - \alpha}(n \smash{ \rsphata{} } + 1,\ n - n \smash{ \rsphata{} })
    \end{equation} 
    provides a one-sided guarantee $\Pr{\rho(\xp) \leq \rsphatmax{\xp}{\alpha}(n, \rsphata{}) } \geq 1 - \alpha$.
    
The cumulative distribution of the Beta distribution is given by:
\[
    \mathrm{F}(x; a,b) = \frac{\mathrm{B}(x; a, b)}{\mathrm{B}(a,b)}
\]
where $\mathrm{B}(x; a, b)$ is the incomplete beta function, defined as:
\[
    \mathrm{B}(x; a, b) = \int_0^x t^{a-1} (1 - t)^{(b-1)} \; dt
\]
and $\mathrm{B}(a,b) = \mathrm{B}(1; a, b)$.

Suppose $\hat{\rho}(\xp) = 0$. Then our parameters for the beta distribution are $a = 1$, $b = n$. Hence,
\[
    \mathrm{F}(x; 1, n) = 1 - (1-x)^n
\]

Since the quantile function is the inverse of the CDF, we have
\[
    \mathsf{B}_{1 - \alpha}(1, n) = 1 - \alpha^{\frac{1}{n}}
\]

To reject $H_0$, we need $\mathsf{B}_{1 - \alpha}(1, n) = 1 - \alpha^{\frac{1}{n}} < \thresh$. By rearranging the inequality, we have
\[
    n > \frac{\ln(\alpha)}{\ln(1-\thresh)}
\]
\end{proof}

\iftoggle{neurips}{
\clearpage
\section*{NeurIPS Paper Checklist}

\begin{enumerate}

\item {\bf Claims}
    \item[] Question: Do the main claims made in the abstract and introduction accurately reflect the paper's contributions and scope?
    \item[] Answer: \answerYes{} %
    \item[] Justification: We include a validation study to validate their properties empirically, and include demonstrations to show their value in practice (\cref{Sec::InferenceGuarantees}, \cref{Sec::Experiments}). We include proofs for all theoretical claims in the supplementary material in \cref{Appendix::OmittedProofs}, and code to reproduce our empirical results at \repourl{}.

\item {\bf Limitations}
    \item[] Question: Does the paper discuss the limitations of the work performed by the authors?
    \item[] Answer: \answerYes{} %
    \item[] Justification: We discuss the overarching limitations of our approach in \cref{Sec::ConcludingRemarks}, and discuss specific limitations with different assumptions or algorithms in the manuscript.

\item {\bf Theory assumptions and proofs}
    \item[] Question: For each theoretical result, does the paper provide the full set of assumptions and a complete (and correct) proof?
    \item[] Answer: \answerYes{} %
    \item[] Justification: We include assumptions and proofs for theoretical claims in the supplementary material under \cref{Appendix::Proofs}.

    \item {\bf Experimental result reproducibility}
    \item[] Question: Does the paper fully disclose all the information needed to reproduce the main experimental results of the paper to the extent that it affects the main claims and/or conclusions of the paper (regardless of whether the code and data are provided or not)?
    \item[] Answer: \answerYes{} %
    \item[] Justification: We include a link to our codebase, which includes all of the code we used to generate the results. We also specify these details in \cref{Sec::Experiments} and \cref{Appendix::ContentModeration}, \cref{Appendix::RecidivismPrediction}.%

\item {\bf Open access to data and code}
    \item[] Question: Does the paper provide open access to the data and code, with sufficient instructions to faithfully reproduce the main experimental results, as described in supplemental material?
    \item[] Answer: \answerYes{} %
    \item[] Justification:
    ICPSR dataset must be requested from ICSPR. We include processing details. We include a link to our GitHub repository, details on our intervention sets, and descriptions of our datasets and experiments in \cref{Sec::Experiments} and \cref{Appendix::ContentModeration,Appendix::RecidivismPrediction}.

\item {\bf Experimental setting/details}
    \item[] Question: Does the paper specify all the training and test details (e.g., data splits, hyperparameters, how they were chosen, type of optimizer) necessary to understand the results?
    \item[] Answer: \answerYes{} %
    \item[] Justification: We describe the setup for all experiments in the main text in \cref{Sec::Experiments}, and provide additional details in  \cref{Appendix::ContentModeration,Appendix::RecidivismPrediction}. We include code to reproduce our results on \repourl{}.

\item {\bf Experiment statistical significance}
    \item[] Question: Does the paper report error bars suitably and correctly defined or other appropriate information about the statistical significance of the experiments?
    \item[] Answer: \answerYes{} %
    \item[] Justification: Yes, we describe the statistical significance, confidence intervals, and include error bars in our figures presented in \cref{Sec::Algorithms,Sec::Experiments} (additional information in \cref{Appendix::ContentModeration}, and \cref{Appendix::RecidivismPrediction}).

\item {\bf Experiments compute resources}
    \item[] Question: For each experiment, does the paper provide sufficient information on the computer resources (type of compute workers, memory, time of execution) needed to reproduce the experiments?
    \item[] Answer: \answerYes{} %
    \item[] Justification: All experiments were run on standard-issue laptops. Audit and reachable set construction time details are in when appropriate (\cref{Appendix::ContentModeration,Appendix::RecidivismPrediction}).

\item {\bf Code of ethics}
    \item[] Question: Does the research conducted in the paper conform, in every respect, with the NeurIPS Code of Ethics \url{https://neurips.cc/public/EthicsGuidelines}?
    \item[] Answer: \answerYes{} %
    \item[] Justification: We confirm that all reviewers have reviewed the code of ethics, and attest that the work conforms to the code.

\item {\bf Broader impacts}
    \item[] Question: Does the paper discuss both potential positive societal impacts and negative societal impacts of the work performed?
    \item[] Answer: \answerYes{} %
    \item[] Justification: We discuss the societal impact of our work in \cref{Sec::Introduction,Sec::Experiments,Sec::ConcludingRemarks}.

\item {\bf Safeguards}
    \item[] Question: Does the paper describe safeguards that have been put in place for responsible release of data or models that have a high risk for misuse (e.g., pre-trained language models, image generators, or scraped datasets)?
    \item[] Answer: \answerNA{} %
    \item[] Justification: Our paper does not pose such risks.

\item {\bf Licenses for existing assets}
    \item[] Question: Are the creators or original owners of assets (e.g., code, data, models), used in the paper, properly credited and are the license and terms of use explicitly mentioned and properly respected?
    \item[] Answer: \answerYes{} %
    \item[] Justification: We wrote our own codebase which lists its license on \repourl{}.

\item {\bf New assets}
    \item[] Question: Are new assets introduced in the paper well documented and is the documentation provided alongside the assets?
    \item[] Answer: \answerYes{} %
    \item[] Justification: We introduce a repository, which includes documentation.

\item {\bf Crowdsourcing and research with human subjects}
    \item[] Question: For crowdsourcing experiments and research with human subjects, does the paper include the full text of instructions given to participants and screenshots, if applicable, as well as details about compensation (if any)?
    \item[] Answer: \answerNA{} %
    \item[] Justification: We did not crowdsource experiments or use human subjects, therefore we do not include these details.

\item {\bf Institutional review board (IRB) approvals or equivalent for research with human subjects}
    \item[] Question: Does the paper describe potential risks incurred by study participants, whether such risks were disclosed to the subjects, and whether Institutional Review Board (IRB) approvals (or an equivalent approval/review based on the requirements of your country or institution) were obtained?
    \item[] Answer: \answerNA{} %
    \item[] Justification: We did not use research with human subjects, thus did not require IRB approval.

\item {\bf Declaration of LLM usage}
    \item[] Question: Does the paper describe the usage of LLMs if it is an important, original, or non-standard component of the core methods in this research? Note that if the LLM is used only for writing, editing, or formatting purposes and does \emph{not} impact the core methodology, scientific rigor, or originality of the research, declaration is not required.
    \item[] Answer: \answerNA{} %
    \item[] Justification: LLMs have nothing to do as important, original or non-standard components and were only used for writing and editing purposes. We evaluate different LLMs in section 4, but only use them in a black box capacity. 

\end{enumerate}

}

\end{document}